\documentclass[conference]{IEEEtran}
\IEEEoverridecommandlockouts
\usepackage{cite}
\usepackage{amsmath,amssymb,amsfonts}
\usepackage{algorithmic}
\usepackage{graphicx}
\usepackage{textcomp}
\usepackage{xcolor}
\usepackage{numprint}
\def\BibTeX{{\rm B\kern-.05em{\sc i\kern-.025em b}\kern-.08em
    T\kern-.1667em\lower.7ex\hbox{E}\kern-.125emX}}

\usepackage{tikz}
\usepackage{pgfplots}
\pgfplotsset{compat=1.17}
\usetikzlibrary{pgfplots.groupplots}
\usepackage{tikz-dependency}
\usetikzlibrary{babel, positioning, matrix, shapes.misc, arrows.meta, backgrounds, fit} 
\pgfdeclarelayer{nodelayer}
\pgfdeclarelayer{edgelayer}
\pgfsetlayers{background,nodelayer,main,edgelayer} 
\usepackage{hyperref}
\usepackage{subcaption} 
\usepackage{cellspace}
\usepackage{amsmath} 
\usepackage{amsfonts} 
\usepackage{amssymb} 
\usepackage{mathptmx}
\usepackage{latexsym}
\usepackage{graphicx} 
\usepackage{booktabs} 
\usepackage[table]{xcolor}
\usepackage{multirow} 
\usepackage{array} 
\usepackage{makecell} 
\usepackage{graphicx} 
\usepackage{tablefootnote}
\usepackage{threeparttable}
\usepackage{float} 
\usepackage{pgf}
\usepackage{pgfplots}
\usepackage[T1]{fontenc}
\usepackage[inline]{enumitem}
\setlist{noitemsep, leftmargin=*}
\usepackage{adjustbox}
\usepackage[american]{babel}
\usepackage[autostyle=true]{csquotes}
\usepackage{expex}

\usepackage{longtable}
\usepackage{enumitem}
\usepackage{graphicx}

\usepackage{pgfplots}
\usepackage{pgfplotstable}
\usepackage{subcaption}
\usepackage{tabularx}
\usepackage{comment}

\usepackage{listings}
\usepackage{cuted}

\lstdefinelanguage{json}{
  basicstyle=\ttfamily\small,
  numbers=left,
  numberstyle=\tiny,
  stepnumber=1,
  numbersep=5pt,
  showstringspaces=false,
  breaklines=true,
  frame=single,
  backgroundcolor=\color{lightgray},
  literate=
   *{0}{{{\color{numb}0}}}{1}
    {1}{{{\color{numb}1}}}{1}
    {2}{{{\color{numb}2}}}{1}
    {3}{{{\color{numb}3}}}{1}
    {4}{{{\color{numb}4}}}{1}
    {5}{{{\color{numb}5}}}{1}
    {6}{{{\color{numb}6}}}{1}
    {7}{{{\color{numb}7}}}{1}
    {8}{{{\color{numb}8}}}{1}
    {9}{{{\color{numb}9}}}{1}
    {:}{{{\color{punct}{:}}}}{1}
    {,}{{{\color{punct}{,}}}}{1}
    {"}{{{\color{delim}{"}}}}{1},
}

\definecolor{numb}{rgb}{0.7,0.2,0.2}
\definecolor{punct}{rgb}{0,0,0}
\definecolor{delim}{rgb}{0,0,0.5}
\definecolor{lightgray}{rgb}{0.95,0.95,0.95}

\usepackage[utf8]{inputenc}

\usepackage{microtype}

\usepackage{inconsolata}
\usepackage{numprint}
\definecolorset{cmyk}{GU-}{}{%
  Goethe-Blau,1,0.2,0,0.4;%
  Hellgrau,0.04,0.04,0.05,0.02;%
  Sandgrau,0.12,0.09,0.13,0;%
  Dunkelgrau,0.25,0.25,0.3,0.75;%
  Purple,0.08,1,0.3,0.36;%
  Emo-Rot,0.04,1,0.8,0.07;%
  Senfgelb,0.01,0.25,1,0.05;%
  Gruen,0.62,0.4,0.87,0.09;%
  Magenta,0.08,0.86,0.12,0.12;%
  Orange,0,0.7,1,0.04;%
  Sonnengelb,0,0.12,0.95,0;%
  Helles-Gruen,0.4,0.17,0.81,0.07;%
  Lichtblau,0.8,0,0.06,0.04}

\definecolor{darkgrey176}{RGB}{176,176,176}
\definecolor{lightgrey204}{RGB}{204,204,204}
\definecolor{salmon}{RGB}{250,128,114}
\definecolor{skyblue}{RGB}{135,206,235}

\newlength{\myeqskip}  
\AtBeginDocument{%
    \setlength\abovedisplayskip{\myeqskip}%
    \setlength\belowdisplayskip{\myeqskip}%
    \setlength\abovedisplayshortskip{\myeqskip-\baselineskip}%
    \setlength\belowdisplayshortskip{\myeqskip}}

\newcommand{\AM}[1]{\textcolor{black}{#1}}
\newcommand{\AMC}[1]{\textcolor{black}{#1}}
\newcommand{\R}[1]{\textcolor{black}{#1}}

\usepackage{xspace}

\definecolor{lightgray}{gray}{0.92}
\newcommand{\LH}[1]{\textcolor{black}{#1}\xspace}

\begin{document}


\title{Learning to Detect Cross-Modal Negation: An Analysis of Latent Representations and an Attention-Based Solution}

\author{\IEEEauthorblockN{1\textsuperscript{st} Ali AbuSaleh*}
\IEEEauthorblockA{\textit{Text Technology Lab} \\ \textit{Department of Computer Science} \\
\textit{Goethe University Frankfurt}\\
Frankfurt, Germany \\
a.abusaleh@em.uni-frankfurt.de}
*Corresponding author
~\\
\and
\IEEEauthorblockN{2\textsuperscript{nd}Leon Hammerla}
\IEEEauthorblockA{\textit{Text Technology Lab}  \\ \textit{Department of Computer Science} \\
\textit{Goethe University Frankfurt}\\
Frankfurt, Germany \\
hammerla@em.uni-frankfurt.de}
~\\
\and
\IEEEauthorblockN{3\textsuperscript{rd} Prof. Dr. Alexander Mehler}
\IEEEauthorblockA{\textit{Text Technology Lab} \\  \textit{Department of Computer Science} \\
\textit{Goethe University Frankfurt}\\
Frankfurt, Germany \\
mehler@em.uni-frankfurt.de}
~\\
}



\maketitle


\begin{center}
    
\textbf{\textcopyright IEEE. This manuscript is an accepted version of the article (ICNLP2026). Published in IEEE Xplore, DOI: \href{https://ieeexplore.ieee.org/abstract/document/11527861}{10.1109/ICNLP69856.2026.11527861} \cite{11527861}: }
\end{center}

\begin{abstract}
    
Detecting high-level semantic concepts like negation across modalities remains a challenge for current multimodal systems.
We analyze this as a fundamental representation learning problem, providing the first evidence that negation does not form a linearly or non-linearly separable class in the latent spaces of standard vision-language models (VLMs).
We demonstrate that pretrained embeddings primarily encode modality-specific features, lacking a generalizable negation signal.
To overcome this, we propose a novel cross-modal attention architecture that explicitly models inter-modal dependencies, achieving performance gains of up to +7.03\% F1 over unimodal baselines. 
Our analysis reveals a key asymmetry: while textual negation often appears independently, visual negation is semantically dependent on linguistic context, 
a finding validated through our statistical analysis of 3,222 political video-text pairs automatically annotated via \textsc{Qwen2.5-VL}.
By combining this analysis with self-supervised video representations (\textsc{JEPA2}), we advance the modeling of temporal negation. 
This work provides new methods and insights for learning robust, semantically-aligned representations in multimodal systems.
\end{abstract}


\section{Introduction}

Negation is central to communication, conveying absence, refusal, or opposition \cite{horn:et:al:2025}. It supports critical thinking and effective debate, with individuals typically relying on speech, visual, or contextual cues to interpret it effortlessly \cite{hasse:2024}.
In contrast, current visual-language models (VLMs) struggle with understanding and generating negation \cite{kumail:et:al:2025}. Given its ubiquity in everyday interaction, this poses a real limitation.
Negation, like affirmation, can cause cognitive dissonance when descriptions conflict with perception \cite{Dudschig:et:al:2017, Dudschig:et:al:2020}. E.g., a politician denying climate change while storm footage suggests otherwise, or drought claims contradicted by visible rainfall.
Such scenarios require careful reasoning across different modes.
Processing negators (e.g., \textit{not}, \textit{never}, or \textit{without}) remains challenging for VLMs, especially when linking them to visuals of absence or opposition. Progress is slow, as subtle context shifts and implied meaning often escape systems focused on direct matches.
To address this challenge, we answer two questions at the intersection of language, vision, and multimodality:

\begin{itemize}
    \item \textbf{Q1:} \textit{
    What visual or non-verbal patterns drive use of negation, and how do visuals influence choosing negation over affirmation?}
    \item \textbf{Q2:} \textit{
    Which visual cues in online media prompt humans to describe scenes in terms of absence, contradiction, or refusal rather than affirmation?}
\end{itemize}
Negation is hard to define because it shifts meaning rather than referring to something tangible \cite{horn:et:al:2025}. It does not add content but reverses the force of existing statements \cite{Matti:2007}.
The abstract nature and variability of negation make it difficult to learn from statistics.
While linguistic negation is usually expressed explicitly, it appears more indirectly in visual media, for example, through contradictions between text and image, or by visually emphasizing empty spaces that suggest absence.
Our analysis reveals three issues in current multimodal setups: (1) negation lacks clear boundaries in typical embeddings, keeping accuracy near chance; (2) single-modality formats capture intramodal but miss crossmodal patterns essential for negation; and (3) simple overlaps fail to align negation across text and image.
To address these issues, we proceed as follow:
We provide a classification of eight negation types, often missed by standard VLMs, arising from text-image relations and enabling finer-grained analysis of meaning shifts when visuals contradict or modify statements.
We use linear discriminant analysis along with wide-ranging classification tests to reveal that negated meanings do not clearly split from positive ones.
We use cross-modal attention to link text and image by filtering irrelevant signals, boosting F1 by up to 7.03\% over single-modality baselines.
Using \textsc{JEPA} \cite{assran:et:al:2025}, we find that deeper video representations reveal negation more clearly than standard image-based encodings.
We use LLMs--specifically Qwen-VL 2.5 \cite{bai:et:al:2025}--to study political discourse.
We analyze \numprint{3222} video-text pairs, revealing fine-grained patterns in how negation manifests across modalities and how these correspond.
We contribute five main advances: (1) an eight-class taxonomy of multimodal negation patterns, (2) diagnostic evidence of limitations in embedding representations, (3) effective cross-modal attention architectures, (4) analysis of specialized video representations (JEPA2), and (5) an LLM-based annotation framework.

\section{Related work}

\textit{Textual Negation Detection:}
\LH{%
Early 
negation detection used rule-based methods or regular expressions (e.g., NegEx, DEEPEN, or NegBio) to identify negation cues and their scope within sentences \cite{Chapman:etal:2001,Mehrabi:etal:2015,Peng:etal:2017}.
%
Supervised models (e.g., CRFs) trained on corpora like BioScope improved cue and scope detection \cite{read:etal:2012, Cruz:etal:2016, Ou:etal:2015}. 
%
More recently, fine-tuned transformers (e.g., NegBERT) outperformed earlier methods on benchmark datasets \cite{Khandelwal:Sawant:2020,Truong:etal:2022,Quian:etal:2024}. 
%
Recent work moves beyond cue–scope models to capture discourse-level negation
\cite{So:etal:2025,vrabcova:etal:2025}.}

\textit{Textual Negation in Twitter-Style Data:}
\LH{Most works on negation in social-media data embed negation handling within sentiment analysis (SA). 
Because negation can invert a statement's sentiment, many Twitter-SA systems apply simple heuristics, static windowing, or preprocessing to flip polarity for negated tokens \cite{Reitan:etal:2015,Gupta:etal:2021}.}
\textit{Depiction of Negation in Visual Media:}
\AMC{\cite{sato:et:al:2023} examine whether 
media like photos and manga can convey negation effectively.
%
In the photo setting, participants captioned images showing absent items (e.g., a trainless station).
%
Humans reliably identified negation using commonsense and background knowledge.
%
In contrast, CNNs performed near chance, likely due to their inability to infer implicit context.
%
While highlighting human strengths with image negation, the study overlooks temporal patterns in video--a gap that we address.}

\textit{Negation Understanding in VLMs:}
\AMC{Building on work with static images, recent studies assess how well modern VLMs handle negation.
%
\cite{kumail:et:al:2025} show that joint VLM embeddings often blur the distinction between affirmations and negations.
%
Using their NegBench benchmark of images and videos, they show that models like CLIP and BLIP perform at chance when distinguishing captions such as `\textit{a cup on the table}' vs.\ `\textit{no cup on the table}'.
%
Fine-tuning on synthetic negated data improved results, pointing to training bias rather than semantic failure.
%
Our work extends this by testing whether latent embeddings encode negation and if cross-modal attention can recover it.}

\textit{Context in Negation Understanding:}
\AMC{The \AM{effect detected} by \cite{kumail:et:al:2025} may stem from limited contextual grounding, not a fundamental failure to model negation \cite{reto:et:al:2022}.
%
Their study shows that with sufficient context, pretrained LMs better distinguish negation from affirmation even without fine-tuning.
%
This is especially relevant in videos, where temporal context supports negation understanding.
%
Our hypothesis--that video negation relies on textual context--aligns with this view and is tested via cross-modal attention.}

\textit{Video-related Architectures for Temporal Reasoning:}
\AMC{Video's temporal structure introduces challenges and opportunities for negation beyond what static images allow.
%
Recent video models like V-JEPA2 \cite{assran:et:al:2025} use self-supervised learning to predict video tokens from context, extending JEPA's semantic representation approach \cite{assran:et:al:2023}.
%
These methods are especially relevant for negation detection, which often relies on temporal reasoning and expectations about absent elements.
%
Reflecting this, we test whether V-JEPA2 embeddings better capture negation than general-purpose VLMs.}

\textit{Cross-Modal Approaches to Negation:}
\AMC{The challenge of negation has spurred interest in cross-modal architectures that leverage complementary, cross-modal signals.
%
X-Poll \cite{gotri:et:al:2022} shows that cross-modal attention significantly improves compositional video-text retrieval.
%
This underscores the value of structured modality alignment, which we extend to political discourse.
%
We extend this with a cross-modal attention framework tailored to the nuanced semantics of negation in political discourse.}

\textit{Semantic Understanding in VLMs:}
Beyond architecture, recent work targets improving semantic understanding in video-language models.
%
\cite{li:et:al:2024} show that adding commonsense and world knowledge boosts performance on tasks involving absence and contradiction.
%
This shift toward semantically aware models enables stronger negation understanding, which we leverage via VLMs like Qwen2.5VL \cite{bai:et:al:2025} for annotation.
%
Our framework applies these semantic advances to negation in political discourse.

\textit{Research Gap and Our Contribution:}
While prior work highlights VLM challenges with negation, most focus on static images or simple video scenarios.
%
We focus on negation in real-world political videos, where multimodal cues interact in complex ways.
%
We combine embeddings, cross-modal attention, and LLM annotation to analyze how negation manifests across modalities in political content.
%
This bridges the gap between controlled benchmarks and real-world negation in VLMs.

\section{Methodology}

\begin{figure}[t]
    \centering
    \includegraphics[width=\linewidth, height=0.8\linewidth]{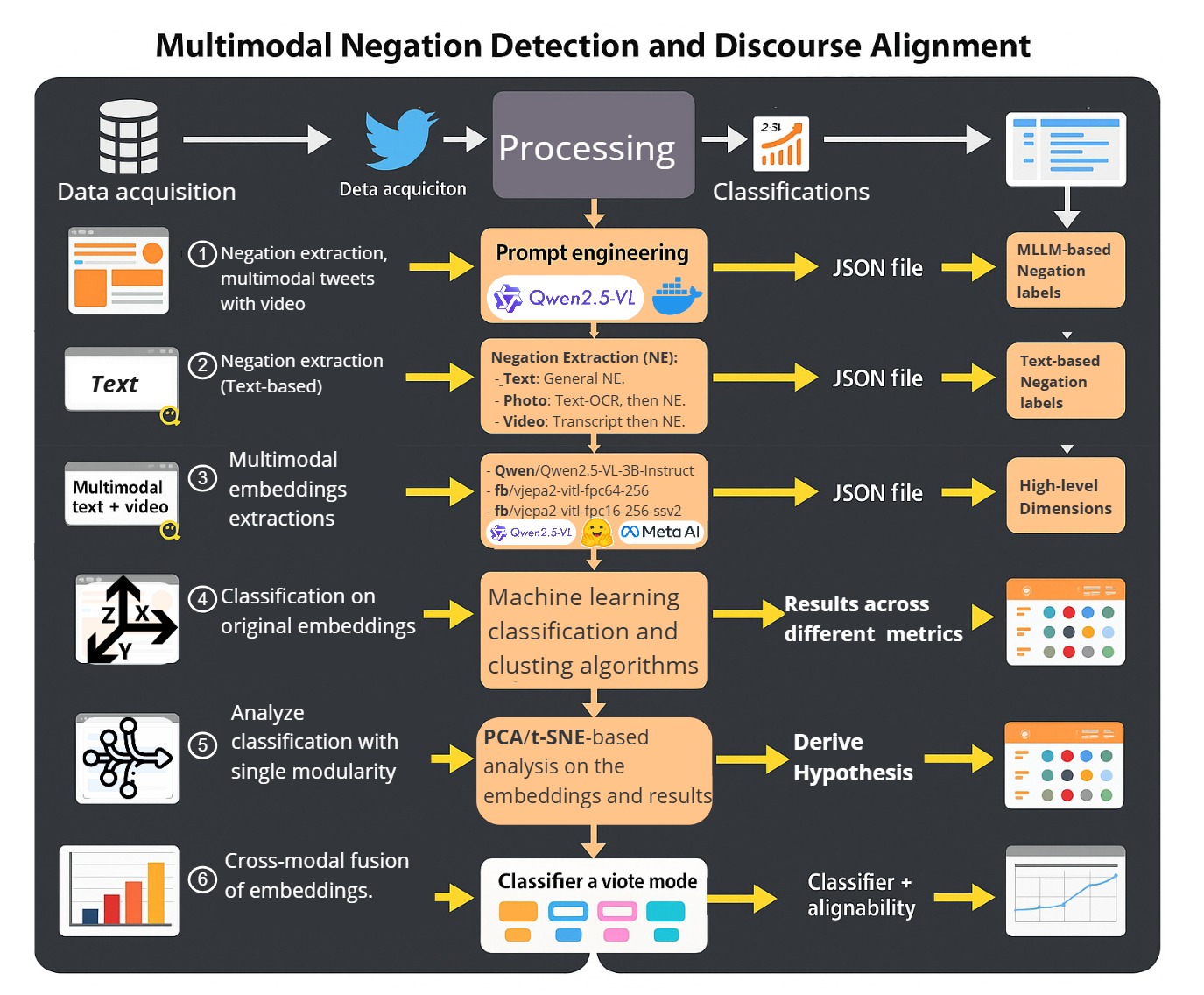}
    \caption{Overall pipeline}
    \label{fig:placeholder}
\end{figure}

Figure~\ref{fig:placeholder} illustrates our overall processing pipeline for detecting and classifying negation patterns in multimodal tweets. 
We begin with a curated dataset of \R{13370 tweets (9359 train, 2006 eval, 2005 test - split)} from  MultiParTweet, \footnote{{Resources could be found in the following \href{https://github.com/aliabusaleh/CrossModalNegation}{Github repository}}}
~\cite{dataset} 
\numprint{3222} containing one media item (video) alongside its accompanying text.
For each tweet, we apply modality-specific transcription: OCR extracts any embedded text from images, while automatic speech recognition processes the audio track of videos. 
R{We then perform negation detection on both the original tweet text and the extracted transcriptions using rule-based and lexical pattern matching. 
}
Next, we extract high-dimensional embeddings from the text and media using multiple pretrained models (Qwen-2.5VL 3B, and JEPA-V2 variants). 
These embeddings are analyzed through a battery of classifiers to assess the separability of negation signals and test our hypothesis about cross-modal representation. 
Finally, we implement cross-modal attention mechanisms to fuse text and video embeddings, enabling joint modeling of negation across modalities.

\subsection{Negation Classes}

\LH{\AMC{
We define eight classes based on possible negation patterns in tweets.
%
Each tweet includes either an image or a video, and all contain one of these media types. 
%
%
%
Classes are defined by the presence or absence of negation in the tweet text and media transcription:
(T\_I, T\_V): Image/video tweet; no negation in text or transcription.
(T\_nI, T\_nV): Image/video tweet; negation only in transcription.
(nT\_I, nT\_V): Image/video tweet; negation only in text.
(nT\_nI, nT\_nV): Image/video tweet; negation in both text and transcription.}}
%

\section{Embeddings analysis}


\subsection{Cluster Analysis LDA}
\LH{\AMC{
We build on the previously defined classes to examine how distinctly they cluster in our data.
%
We first apply Linear Discriminant Analysis (LDA) to project the data into class-dimensional space, then reduce it to two dimensions with t-SNE for visualization.
%
We perform this analysis on three views of the tweet data: (1) tweet text embedding, (2) media embedding, and (3) transcribed text embedding from accompanying images or videos.
%
We generate nine plots: one for all tweets combined, and separate ones for image-only and video-only tweets.
%
All plots show class-based clustering, with the clearest separation in video tweets, where all three views yield well-defined clusters.
%
Class separation is weaker in the combined and image-only sets, though slightly clearer in the latter.
%
Transcribed text embeddings consistently show the best class separation, outperforming tweet-text and media embeddings.
%
The results suggest that transcribed visual content captures semantic cues especially effective for class separation, particularly in video tweets.}}
\begin{figure}[t]
\centering
\begin{subcaptionblock}{\linewidth}
\centering
\includegraphics[width=\linewidth]{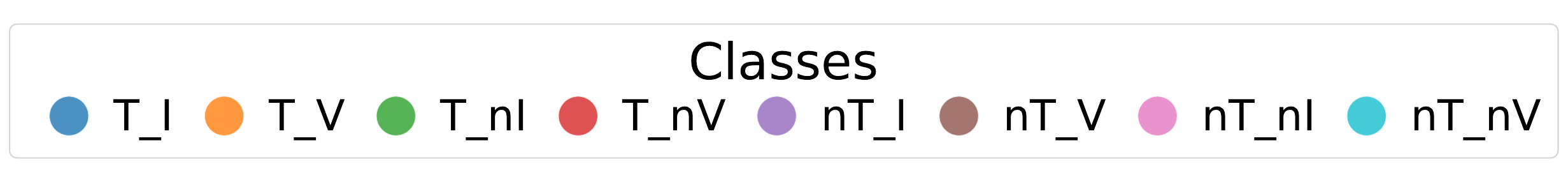}
\end{subcaptionblock}

\begin{subcaptionblock}{\linewidth}
\centering
\begin{subfigure}{0.32\linewidth}
    \includegraphics[width=\linewidth]{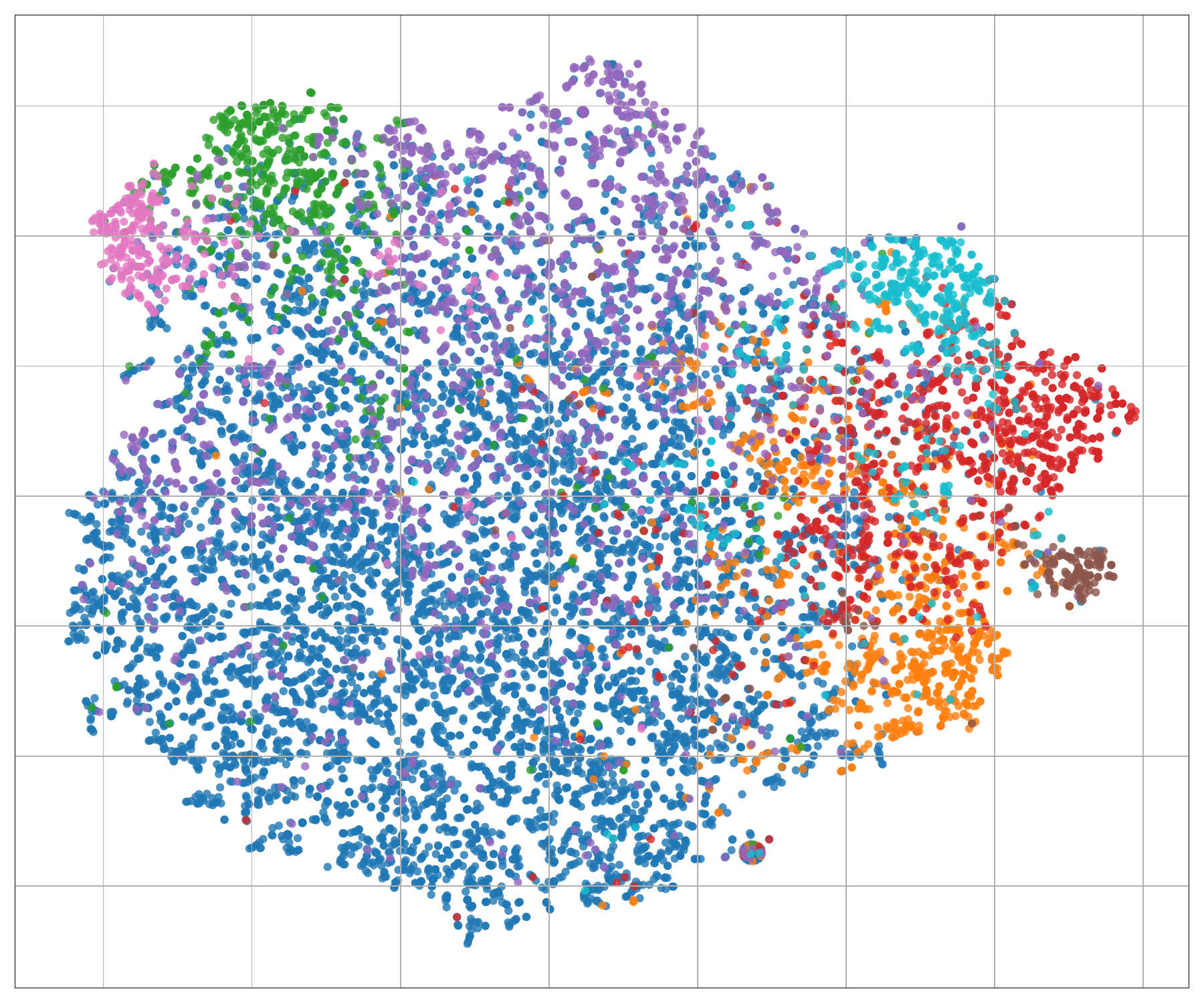}
    \caption{Media}
\end{subfigure}\hfill
\begin{subfigure}{0.32\linewidth}
    \includegraphics[width=\linewidth]{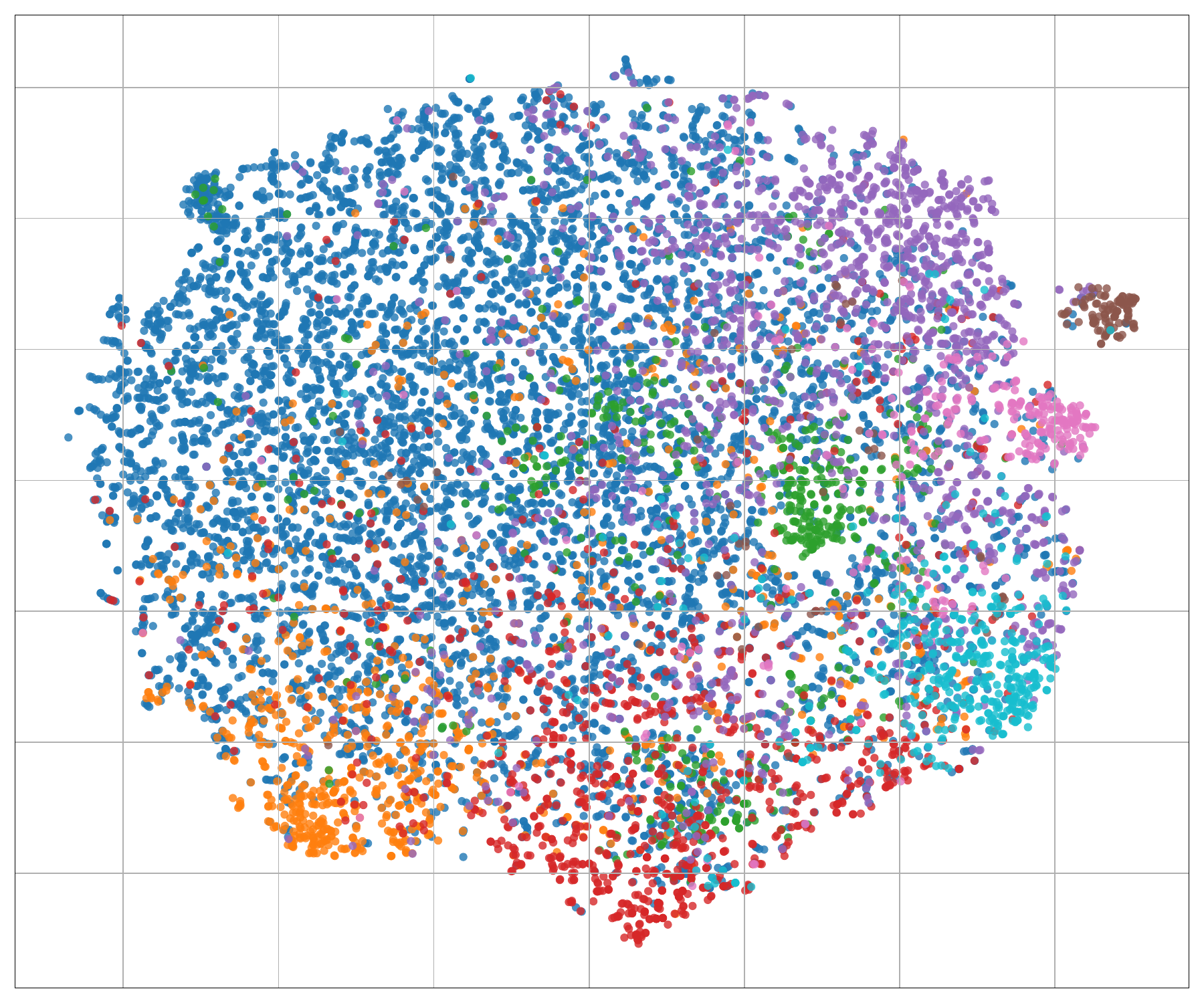}
    \caption{Text}
\end{subfigure}\hfill
\begin{subfigure}{0.32\linewidth}
    \includegraphics[width=\linewidth]{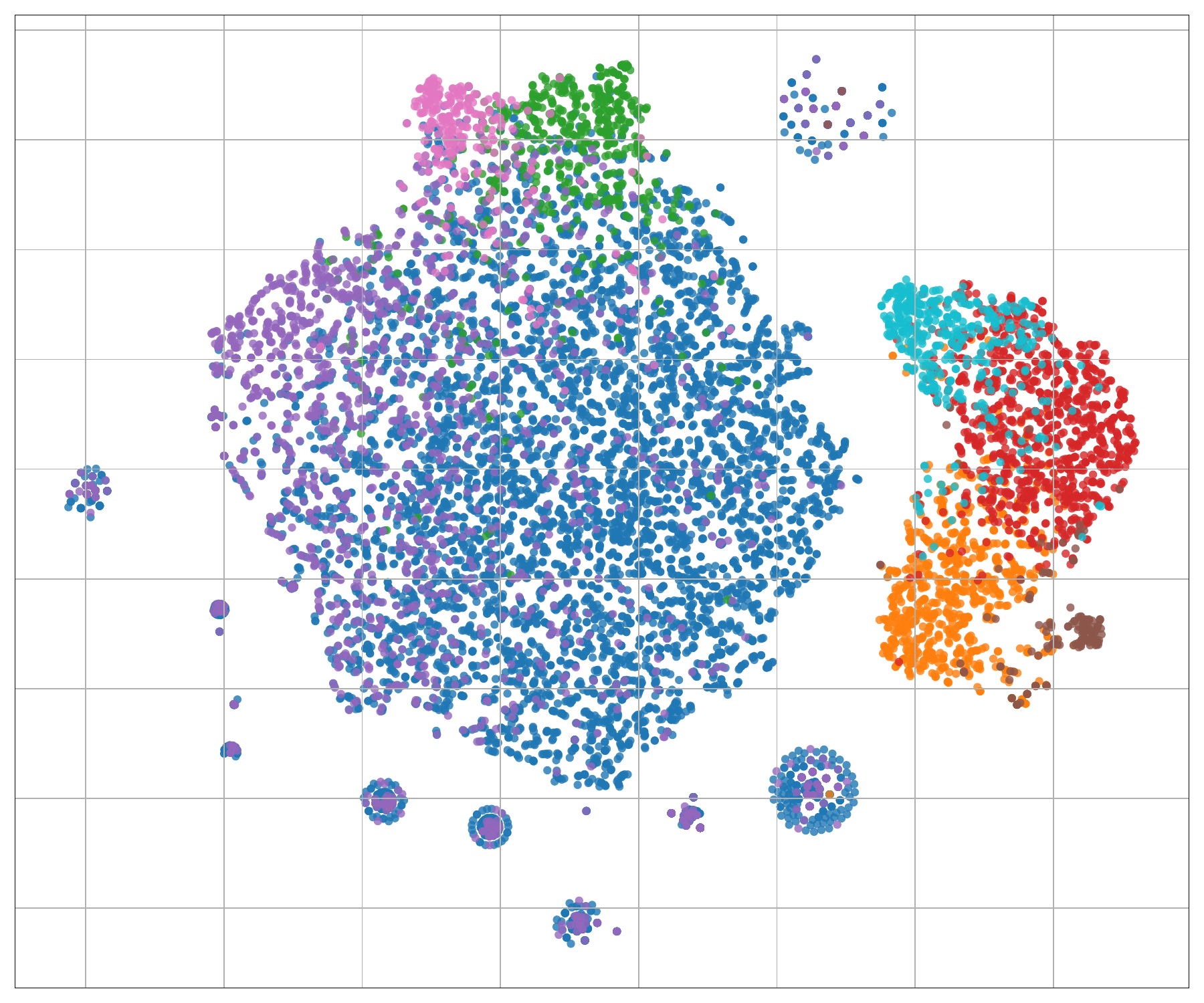}
    \caption{Transcript}
\end{subfigure}
\caption{\textbf{Embeddings from tweets with images and videos}}
\label{fig:proj_images_videos}
\end{subcaptionblock}

\vspace{0.3cm} 

\begin{subcaptionblock}{\linewidth}
\centering
\begin{subfigure}{0.32\textwidth}
    \includegraphics[width=\linewidth]{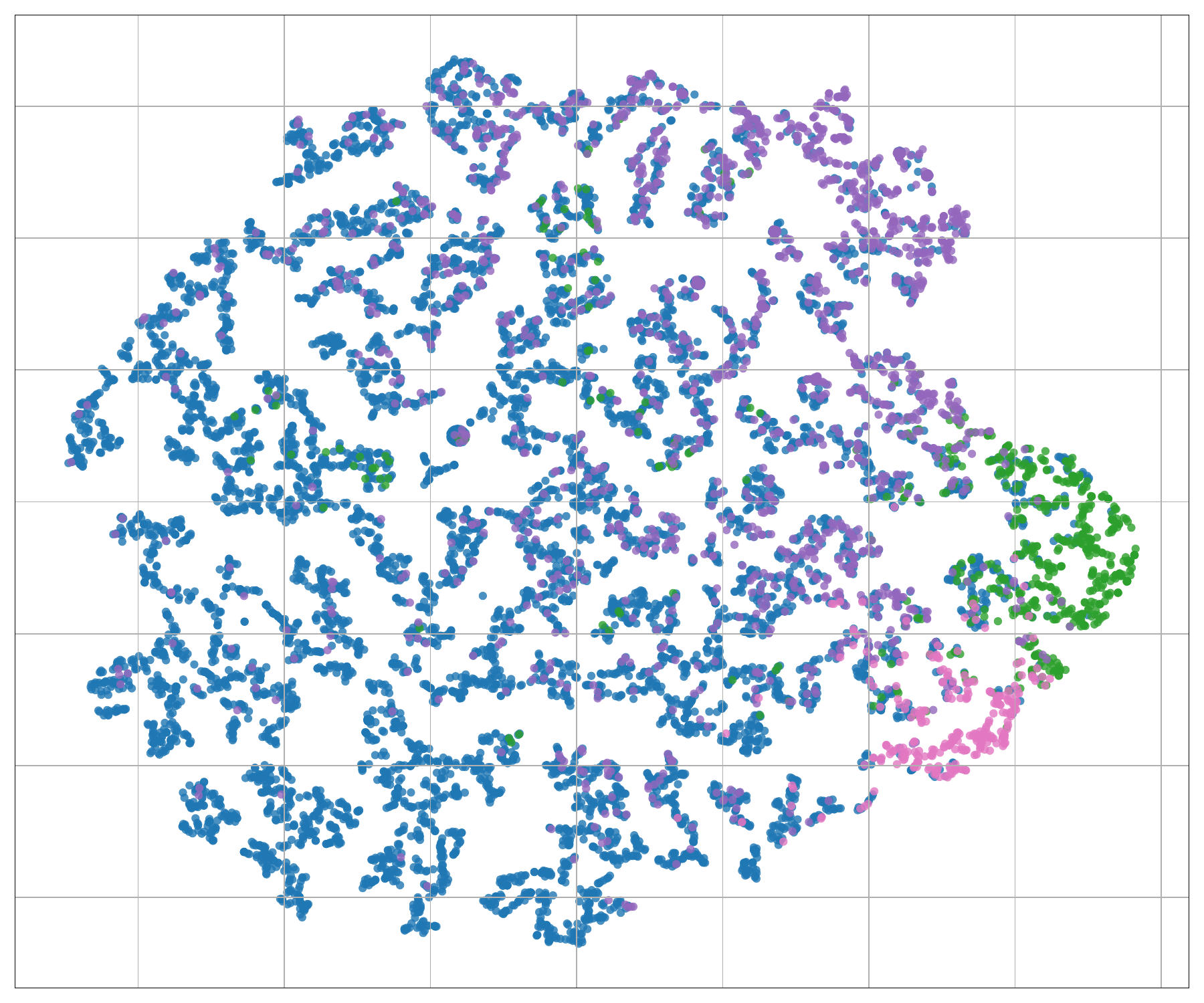}
    \caption{Media}
\end{subfigure}\hfill
\begin{subfigure}{0.32\textwidth}
    \includegraphics[width=\linewidth]{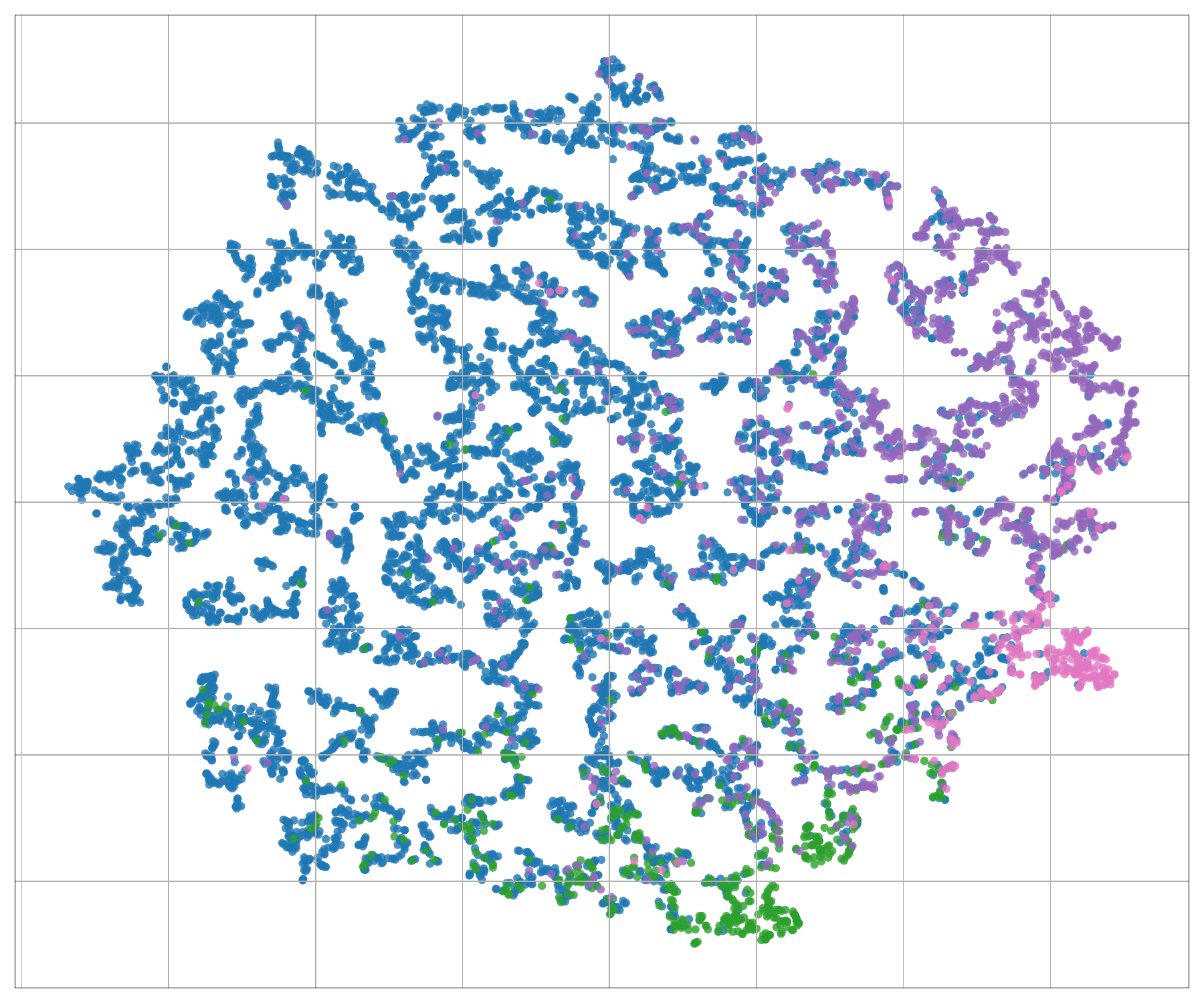}
    \caption{Text}
\end{subfigure}\hfill
\begin{subfigure}{0.32\textwidth}
    \includegraphics[width=\linewidth]{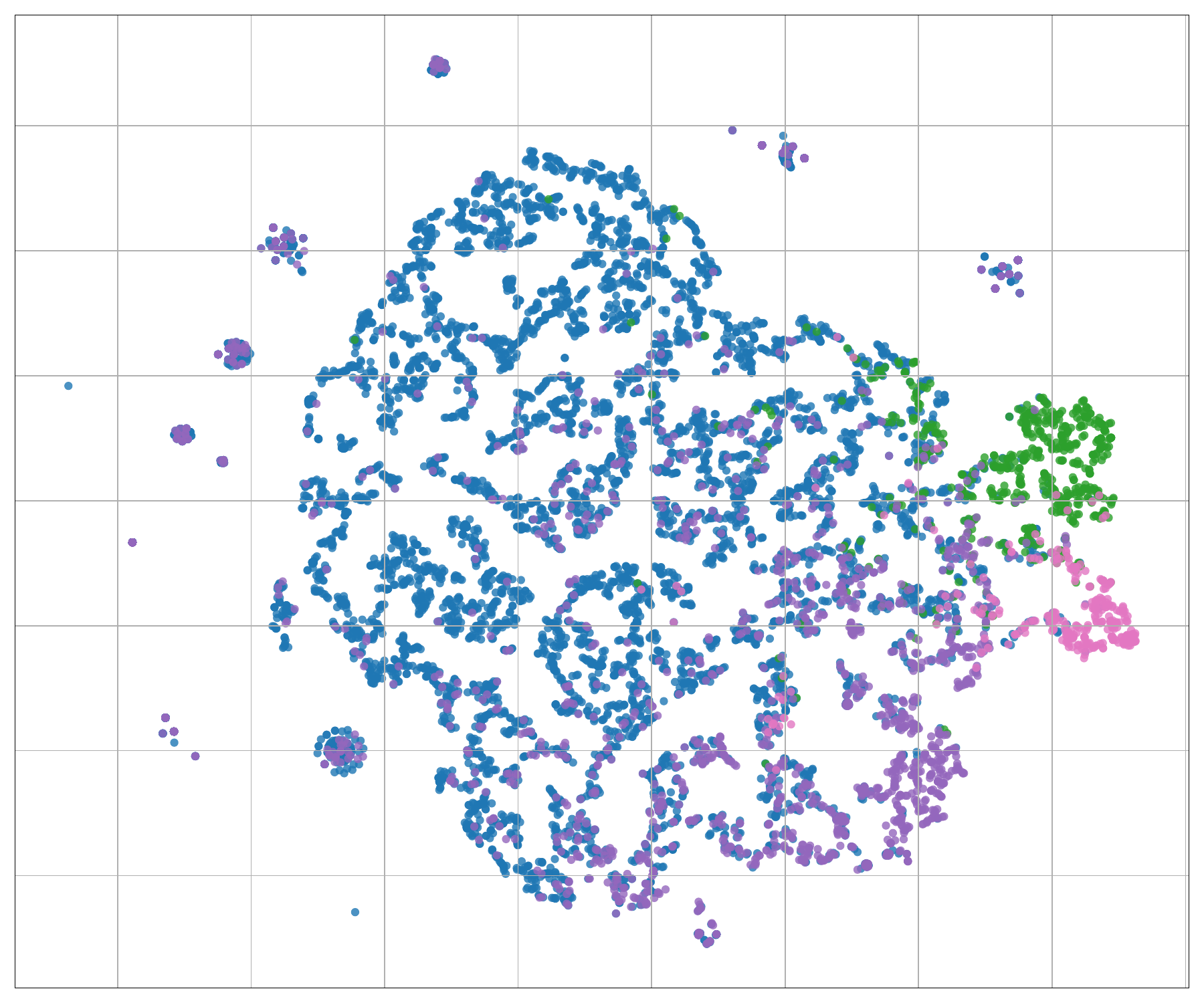}
    \caption{Transcript}
\end{subfigure}
\caption{\textbf{Embeddings from tweets with images only}}
\label{fig:proj_images}
\end{subcaptionblock}

\vspace{0.3cm} 

\begin{subcaptionblock}{\linewidth}
\centering
\begin{subfigure}{0.32\linewidth}
    \includegraphics[width=\linewidth]{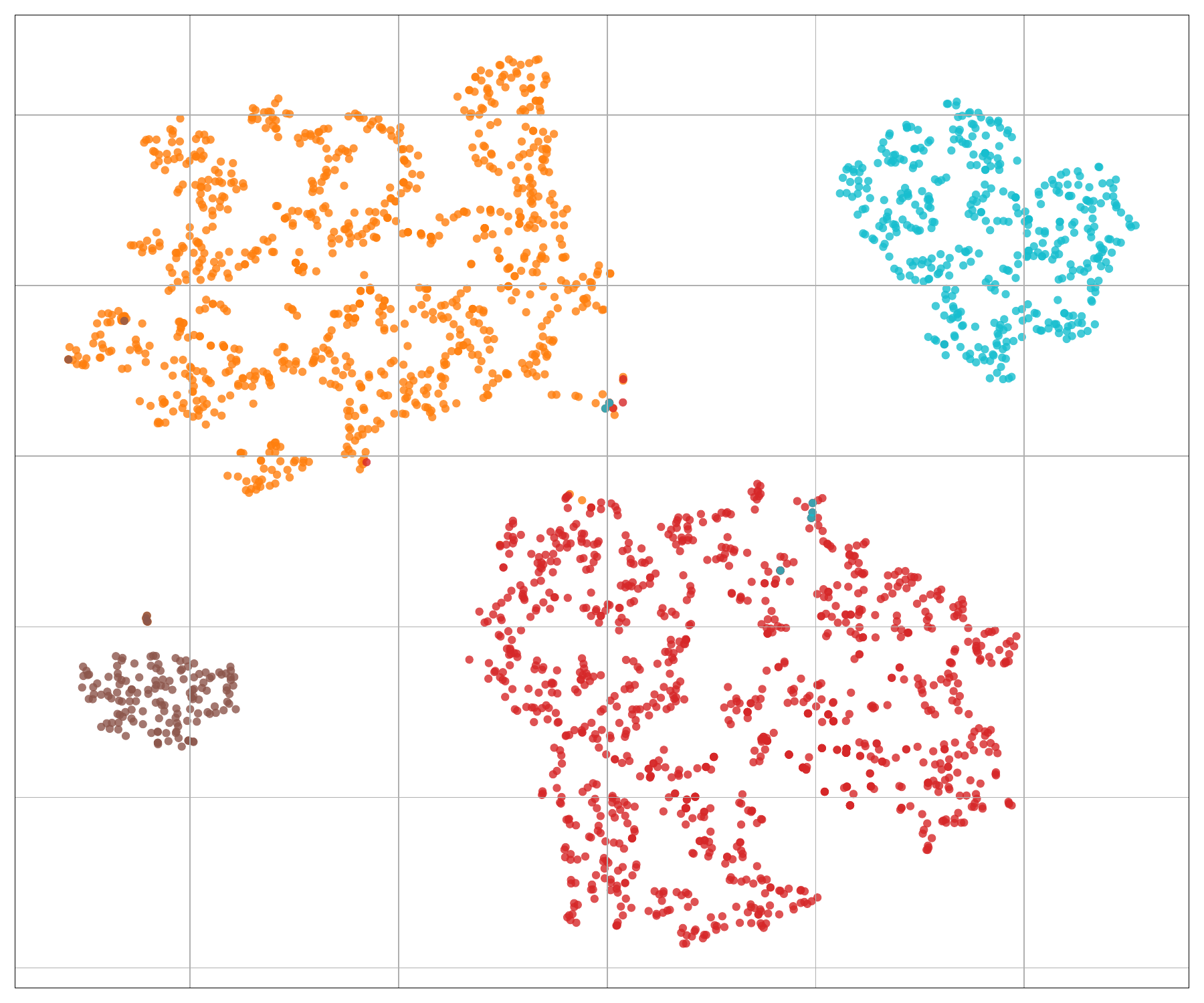}
    \caption{Media}
\end{subfigure}\hfill
\begin{subfigure}{0.32\linewidth}
    \includegraphics[width=\linewidth]{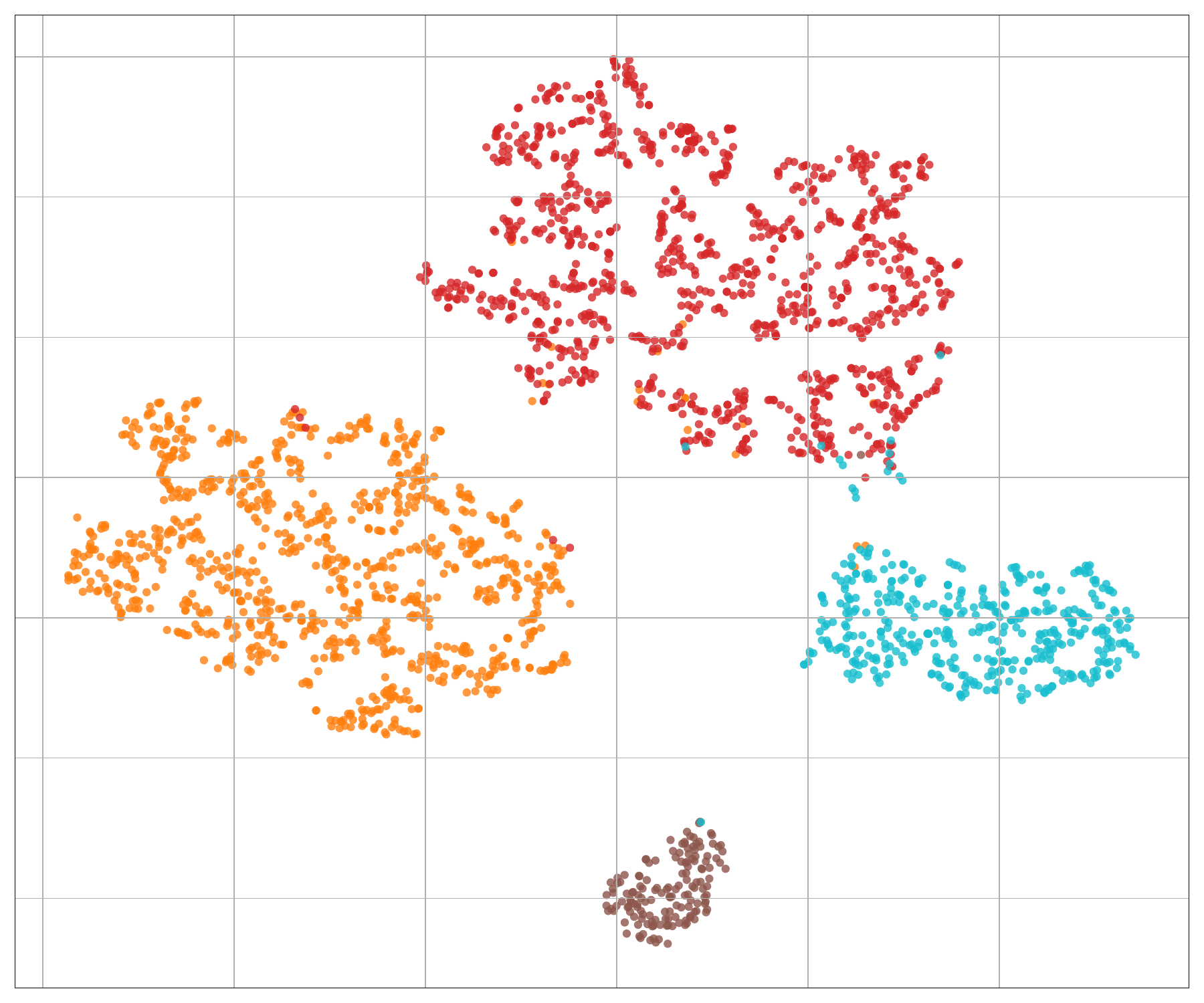}
    \caption{Text}
\end{subfigure}\hfill
\begin{subfigure}{0.32\linewidth}
    \includegraphics[width=\linewidth]{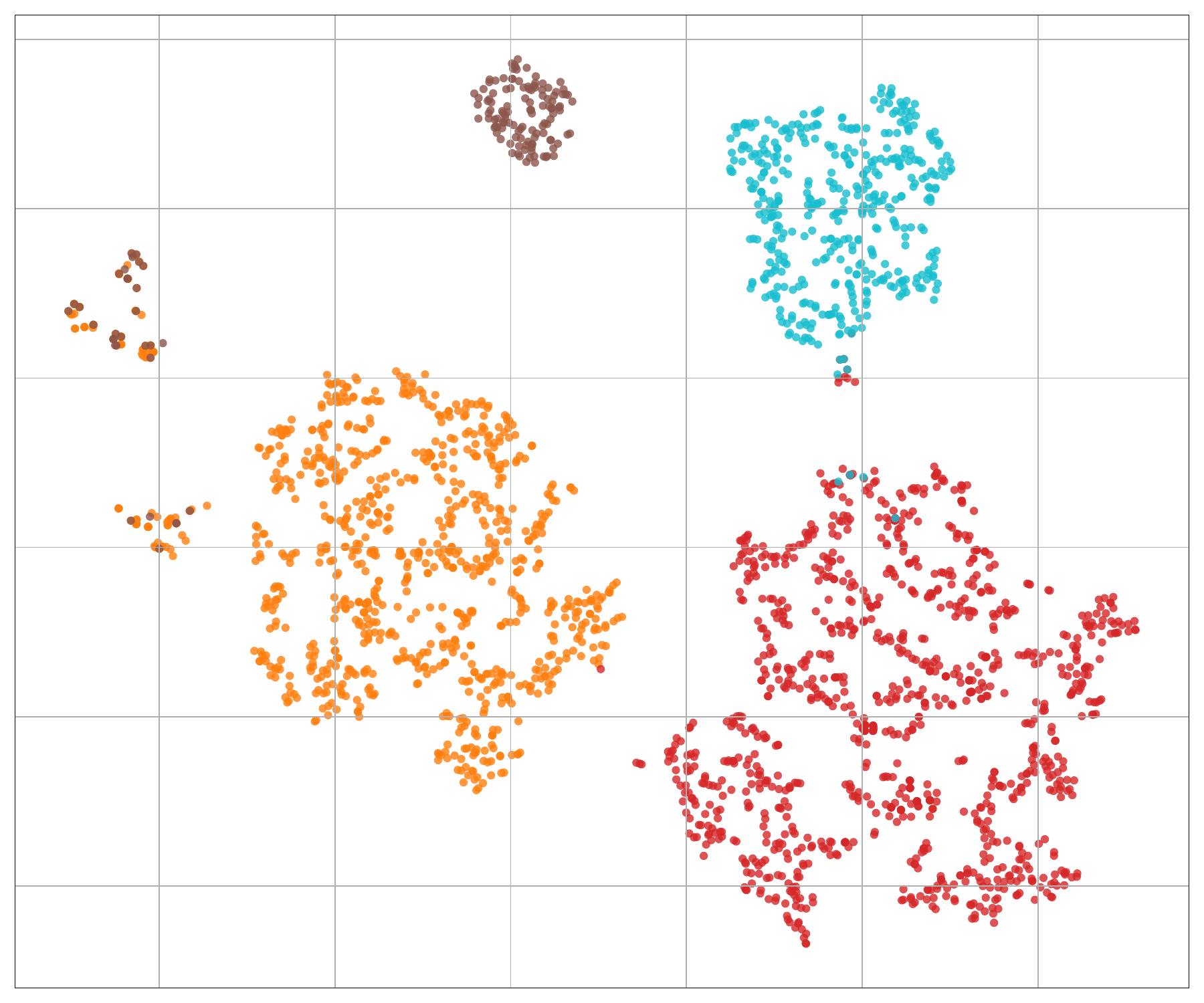}
    \caption{Transcript}
\end{subfigure}
\caption{\textbf{Embeddings from tweets with videos only}}
\label{fig:proj_videos}
\end{subcaptionblock}

\caption{t-SNE visualization of LDA features for media, tweet text, and transcribed text embeddings.}
\end{figure}

\subsection{Text-Label, Media-Embeddings Classifications}
\label{subsection:embeddings_text_label}
\AMC{We test whether text-based negation labels can be mapped to media-only embeddings from Qwen-2.5VL to capture latent negation signals.
%
We assigned text-based negation labels to video, image, and text embeddings, and tested various classifiers to separate negated from non-negated samples.}
%
\AMC{We use diverse classifiers across algorithmic paradigms (Table~\ref{tab:negation_perf}), all with scikit-learn defaults for baseline comparison.}

\paragraph{Classification Results} 
Table~\ref{tab:negation_perf} presents classification performance across the full set of models for negation detection using media embeddings. 
%
Although several classifiers achieve superficially high accuracy (up to approximately 81\%), Balanced Accuracy and ROC AUC scores remain close to 0.5 across nearly all models, indicating performance at chance level.
Notably, simple baselines such as the DummyClassifier perform comparably to more sophisticated algorithms, suggesting that the latent negation signal in these embeddings is weak or not linearly/non-linearly separable.

\paragraph{Embedding Visualization via PCA} 
To understand why classifiers performed at chance, we applied Principal Component Analysis (PCA) to the media embeddings and visualized the first 50 components in Figure~\ref{fig:parallel_coordinates}. 
PCA allows the reduction of high-dimensional embeddings into a lower-dimensional space that preserves the directions of greatest variance, facilitating the identification of systematic patterns that may be masked in the original high-dimensional space. 

The PCA visualizations revealed systematic differences by modality-embedding, indicating that embeddings primarily encode modality-specific features rather than negation.
This observation motivated additional experiments in which we varied both the input embeddings and the label sources across modalities. 
For instance, an image or video was assigned the negation label if a negation cue was identified within its OCR-derived text or its associated transcription data

\paragraph{Modality-Label Experiments} 
Table~\ref{tab:rnn-model-metrics} presents RNN-based model performance across combinations of embedding modalities and label sources. 
The results indicate that classifier performance is heavily influenced by the alignment between input embeddings and labels. Despite these variations, no configuration clearly isolates a latent negation signal. 
These findings provide the foundation for our subsequent cross-modal experiments, where multiple modalities are integrated to investigate whether joint representations can better capture negation.
\begin{table*}[t]
\centering
\caption{RNN-Based Model performance grouped by modality category (video, text, photo) and training epoch. Checkmarks indicate presence of each modality as input or label.}
\resizebox{\textwidth}{!}{%
\begin{tabular}{|c|l|c|c|c|ccc|ccc|}
\hline
\textbf{Epoch} & \textbf{Model} & \textbf{Accuracy} & \textbf{F1} & \textbf{ROC AUC} & \multicolumn{3}{c|}{\textbf{Inputs (embeddings)}} & \multicolumn{3}{c|}{\textbf{Labels}} \\
\cline{6-11}
 & & & & & \textbf{Text} & \textbf{Video} & \textbf{Photo} & \textbf{Text} & \textbf{Video} & \textbf{Photo} \\
\hline
\multicolumn{11}{|c|}{\textbf{Video-based Embeddings}} \\
\hline
\multirow{2}{*}{30}
& Video Model         & 0.6957 & 0.6605 & 0.7162 &  & \checkmark &  &  & \checkmark &  \\
& Video-Text Model    & 0.6916 & 0.6359 & 0.5718 &  & \checkmark &  & \checkmark &  &  \\
\cline{1-11}
\multirow{2}{*}{50}
& Video Model         & 0.6489 & 0.6209 & 0.6948 &  & \checkmark &  &  & \checkmark &  \\
& Video-Text Model    & 0.6932 & 0.6331 & 0.5707 &  & \checkmark &  & \checkmark &  &  \\
\hline
\multicolumn{11}{|c|}{\textbf{Text-based Embeddings}} \\
\hline
\multirow{4}{*}{30}
& \makecell{Text Model   \tiny (video-contains)}   & 0.8037 & 0.7883 & 0.8569 & \checkmark &  &  & \checkmark &  &  \\
& Text-Video Model    & 0.7614 & 0.7408 & 0.7758 & \checkmark &  &  &  & \checkmark &  \\
& \makecell{Text Model   \tiny (photo-contains)}   & 0.8499 & 0.8135 & 0.9040 & \checkmark &  &  & \checkmark &  &  \\
& Text-photo Model    & 0.9327 & 0.9010 & 0.7278 & \checkmark &  &  &  &  & \checkmark \\
\cline{1-11}
\multirow{4}{*}{50}
& \makecell{Text Model   \tiny (video-contains)}  & 0.8045 & 0.7995 & 0.8574 & \checkmark &  &  & \checkmark &  &  \\
& Text-Video Model    & 0.7384 & 0.6968 & 0.7679 & \checkmark &  &  &  & \checkmark &  \\
&  \makecell{Text Model   \tiny (photo-contains)}   & 0.8597 & 0.8323 & 0.9090 & \checkmark &  &  & \checkmark &  &  \\
& Text-photo Model    & 0.9310 & 0.9010 & 0.7313 & \checkmark &  &  &  &  & \checkmark \\
\hline
\multicolumn{11}{|c|}{\textbf{Photo-based Embeddings}} \\
\hline
\multirow{2}{*}{30}
& Photo Model         & \textbf{0.9441} & 0.9333 & 0.8632 &  &  & \checkmark &  &  & \checkmark \\
& Photo-text Model    & 0.7837 & 0.7365 & 0.5891 &  &  & \checkmark & \checkmark &  &  \\
\cline{1-11}
\multirow{2}{*}{50}
& Photo Model         & 0.9421 & 0.9316 & 0.8536 &  &  & \checkmark &  &  & \checkmark \\
& Photo-text Model    & 0.7887 & 0.7384 & 0.5907 &  &  & \checkmark & \checkmark &  &  \\
\hline
\end{tabular} }

\label{tab:rnn-model-metrics}
\end{table*}

\subsection{Cross-Modal Attention Framework}


We implement cross modal attention to integrate text and video embeddings
$\mathbf{V} \in \mathbb{R}^{B \times d}$ and
$\mathbf{T} \in \mathbb{R}^{B \times d}$ within a unified transformer architecture.
The model concatenates the embeddings as
\begin{equation}
\mathbf{X} = \mathrm{Concat}(\mathbf{V}, \mathbf{T}),
\label{eq:concat}
\end{equation}
and applies multi head attention,
\begin{equation}
\mathrm{MultiHead}(\mathbf{X})
= \mathrm{Concat}(\mathrm{head}_1, \dots, \mathrm{head}_h)\mathbf{W}^O,
\label{eq:multihead}
\end{equation}
where
\begin{equation}
\mathrm{head}_i
= \mathrm{Attention}(\mathbf{X}\mathbf{W}^Q_i,
\mathbf{X}\mathbf{W}^K_i,
\mathbf{X}\mathbf{W}^V_i).
\label{eq:head}
\end{equation}

\subsubsection{Comparative Framework}
\begin{itemize}
    \item \textbf{Cross-modal}: Joint processing of $\mathbf{V}$ and $\mathbf{T}$
    \item \textbf{Self-attention (unimodal)}: Isolated processing of $\mathbf{V}$ or $\mathbf{T}$
\end{itemize}

\subsubsection{Cross-Modal Advantages}
Cross-attention outperforms unimodal self-attention by modeling intermodal dependencies that support contextual disambiguation (e.g., resolving textual sarcasm via visual cues), crossmodal feature alignment, and complementary signal fusion.
The 4-head cross-attention text-sentiment model achieves superior performance (Acc: 0.8342, F1: 0.8285) versus unimodal text processing (Acc: 0.7889, F1: 0.7582), representing +4.53\% accuracy and +7.03\% F1 gains.

\subsubsection{Head Configuration Analysis}
Optimal performance occurs with 4 attention heads, outperforming 8-head configurations (e.g., +2.52\% accuracy for text-sentiment). 
This suggests reduced heads mitigate overparameterization in embedding spaces of moderate dimensionality.

\begin{table}[t]
\centering
\caption{Performance of attention architectures (30 epochs). CM: Cross-modal, SA: Self-attention.}
\label{tab:attn_results}
\resizebox{\linewidth}{!}{
\begin{tabular}{@{}cccccclc@{}}
\toprule
\multirow{2}{*}{\textbf{Heads}} & \multirow{2}{*}{\textbf{Type}} & \multicolumn{3}{c}{\textbf{Metrics}} & \multicolumn{2}{c}{\textbf{Input}} & \multirow{2}{*}{\textbf{\makecell{Target \\ sentiment}}} \\
\cmidrule(lr){3-5} \cmidrule(lr){6-7}
 & & Acc. & F1 & ROC-AUC & Text & Video & \\ 
\midrule
\multicolumn{8}{c}{\textit{Cross-Modal Attention}} \\
\midrule
8 & CM-Video & 0.7705 & 0.7721 & 0.8964 & \checkmark & \checkmark & Video-Neg \\
8 & CM-Text & 0.8090 & 0.8040 & 0.8916 & \checkmark & \checkmark & Text-Neg\\
4 & CM-Video & 0.7672 & 0.7605 & 0.9038 & \checkmark & \checkmark & Video-Neg \\
4 & CM-Text & \textbf{0.8342} & \textbf{0.8285} & \textbf{0.9136} & \checkmark & \checkmark & Text-Neg \\
\midrule
\multicolumn{8}{c}{\textit{Self-Attention}} \\
\midrule
8 & SA-Video & 0.6499 & 0.6374 & 0.7574 & & \checkmark & Video-Neg \\
8 & SA-Text & 0.7755 & 0.7472 & 0.9051 & \checkmark & & Text-Neg \\
4 & SA-Video & 0.6415 & 0.6305 & 0.7541 & & \checkmark & Video-Neg \\
4 & SA-Text & 0.7889 & 0.7582 & 0.9037 & \checkmark & & Text-Neg \\
\bottomrule
\end{tabular}}
\vspace{0.2cm}
\\ \footnotesize \textit{Note}: Photo modality excluded as unused in experiments
\end{table}

\subsection{JEPA-V2-based embeddings}

We extend our analysis to specialized video representations using two variants of the JEPA-V2 model: base JEPA-V2~\cite{assran:et:al:2023} and JEPA-V2 pretrained on the Something-Something-V2 (SSV)~\cite{goyal:et:al:2017:somethingsomethingvideodatabase} human action dataset. 
These models provide high-level video features that capture temporal and action-oriented information, which we hypothesize may better encode visual negation cues. 
By comparing their performance with standard VLM embeddings (Qwen), we assess how video-specific pretraining affects cross-modal attention mechanisms for negation detection.
\subsubsection{Impact of High-Level Representations on Attention Mechanisms}

Our experiments examine how attention head configuration interacts with high-level embeddings from different models. 
%
As shown in Table~\ref{tab:attn_results}, 4-head cross-modal attention with Qwen embeddings achieves the best performance (accuracy: \numprint{0.8342}, F1: \numprint{0.8285}) for text-negation detection, outperforming 8-head configurations. 
The specialized video representations from JEPA models show distinct performance patterns, with JEPA-V2-SSV demonstrating improved alignment capabilities. 
This suggests that in moderate-dimensional embedding spaces, fewer attention heads mitigate overparameterization while still effectively modeling cross-modal interactions, and that the choice of base embedding model significantly impacts attention mechanism effectiveness for negation detection.
\begin{table}[t]
\centering
\caption{Performance of attention types (30 epochs, 4 heads). CM: Cross-modal, SA: Self-attention.}
\label{tab:attn_results_new}
\resizebox{\linewidth}{!}{
\begin{tabular}{@{}cccccclc@{}}
\toprule
\multirow{2}{*}{\textbf{\makecell{ Attention \\ Heads }}} & \multirow{2}{*}{\textbf{Type}} & \multicolumn{3}{c}{\textbf{Metrics}} & \multicolumn{2}{c}{\textbf{Input}} & \multirow{2}{*}{\textbf{\makecell{Target \\ sentiment}}} \\
\cmidrule(lr){3-5} \cmidrule(lr){6-7}
 & & Acc. & F1 & ROC-AUC & Text & Video & \\ 
\midrule
\multicolumn{8}{c}{\textit{Cross-Modal Attention (V-JEPA2)}} \\
\midrule
4 & CM-Video & 0.7513 & 0.7517 & 0.8100 & \checkmark & \checkmark & Video-Neg \\
4 & CM-Text  & 0.7513 & 0.7439 & 0.7891 & \checkmark & \checkmark & Text-Neg \\
\midrule
\multicolumn{8}{c}{\textit{Cross-Modal Attention (V-JEPA2 SSV)}} \\
\midrule
4 & CM-Video & 0.7513 & 0.7472 & 0.8019 & \checkmark & \checkmark & Video-Neg \\
4 & CM-Text  & 0.8016 & 0.8004 & 0.8290 & \checkmark & \checkmark & Text-Neg \\
\midrule
\multicolumn{8}{c}{\textit{Self-Attention(V-JEPA2)}} \\
\midrule
4 & SA-Video & 0.7460 & 0.7469 & 0.8296 & & \checkmark & Video-Neg \\
4 & SA-Text  & 0.8016 & 0.7837 & 0.8204 & \checkmark & & Text-Neg \\
\midrule
\multicolumn{8}{c}{\textit{Self-Attention  (V-JEPA2 SSV)}} \\
\midrule
4 & SA-Video & 0.6825 & 0.6573 & 0.7557 & & \checkmark & Video-Neg \\
\midrule
\bottomrule
\end{tabular}}
\vspace{0.2cm}
\end{table}
\subsection{Attention Weight Distribution Analysis}

To understand how cross-modal attention processes negation, we analyze attention distributions for models trained on text negation and video negation targets (Table~\ref{tab:attention_comparison}).

\begin{table}[ht]
\centering
\caption{Attention patterns by prediction target. CM = Cross-modal total (Video$\rightarrow$Text + Text$\rightarrow$Video).}
\label{tab:attention_comparison}
\begin{tabular}{lccccc}
\toprule
\textbf{Target} & \textbf{V$\rightarrow$V} & \textbf{V$\rightarrow$T} & \textbf{T$\rightarrow$V} & \textbf{T$\rightarrow$T} & \textbf{CM} \\
\midrule
Text Negation   & 0.542 & 0.458 & 0.449 & 0.551 & 0.907 \\
Video Negation  & 0.258 & 0.742 & 0.642 & 0.358 & 1.383 \\
\bottomrule
\end{tabular}
\end{table}

The two models exhibit fundamentally different strategies.
For text negation, cross-modal attention is asymmetric (90.7\% total) and, critically, differs significantly between negated and non-negated samples: non-negated samples show high cross-modal attention ($\mu=0.973$), while negated samples show a 29.4\% decrease ($\mu=0.678$, $p<0.0001$). 
This suggests the model suppresses alignment when modalities contradict.

For video negation, cross-modal attention is substantially higher (138.3\%) and nearly symmetric, with both modalities strongly attending to each other. 
However, we find no difference between negated and non-negated samples ($p=0.817$). This reveals that visual negation requires constant cross-modal grounding--the model maintains high alignment regardless of negation status, consistent with our finding that visual negation depends on linguistic context.

\subsection{Cross-Modal Framework with FLAVA Embeddings}
\label{sec:flava_comparison}

\begin{table}[t]
\centering
\caption{Cross-modal attention vs. unimodal baselines with FLAVA embeddings. \protect \footnotemark }
\label{tab:flava_crossmodal}
\resizebox{0.9\linewidth}{!}{
\begin{tabular}{lccccl}
\toprule
\textbf{Model} & \textbf{Accuracy} & \textbf{F1} & \textbf{ROC AUC} & \textbf{Input} & \textbf{Target} \\
\midrule
\multicolumn{6}{c}{\textit{Text Negation}} \\
\midrule
Cross-modal & \textbf{0.8022} & \textbf{0.8022} & 0.8546 & Text+Video & Text-Neg \\
Text-only   & 0.7363 & 0.7021 & \textbf{0.8591} & Text & Text-Neg \\
Video-only  & 0.5275 & 0.5023 & 0.6560 & Video & Text-Neg \\
\midrule
\multicolumn{6}{c}{\textit{Video Negation}} \\
\midrule
Cross-modal & \textbf{0.7912} & \textbf{0.7907} & \textbf{0.8546} & Text+Video & Video-Neg \\
Text-only   & 0.7253 & 0.7239 & 0.8130 & Text & Video-Neg \\
Video-only  & 0.5165 & 0.4754 & 0.6386 & Video & Video-Neg \\
\bottomrule
\end{tabular}}
\end{table}

\footnotetext{The experiments is conducted on a subset of the data with a total of 527 randomly selected samples}

To isolate the benefit of cross-modal attention independent of specific feature extractors, we conducted experiments using FLAVA embeddings\cite{singh2022flavafoundationallanguagevision} for both text and video. 
Table~\ref{tab:flava_crossmodal} compares unimodal classifiers against our cross-modal attention framework using identical features.
For text negation, cross-modal attention achieves 80.22\% accuracy, outperforming text-only (73.63\%, +6.6\% gain) and video-only (52.75\%, +27.5\% gain). 
For video negation, cross-modal attention reaches 79.12\% accuracy, exceeding text-only (72.53\%, +6.6\% gain) and video-only (51.65\%, +27.5\% gain).
F1 scores show similar improvements: +10.0\% over text-only and +30.0\% over video-only for text negation; +6.7\% and +31.5\% respectively for video negation.
These results confirm that \textbf{explicitly modeling inter-modal dependencies}--not just better feature extractors--drives our framework's effectiveness. 
The consistent cross-modal advantage across both targets demonstrates that fusion itself, independent of embedding choice, is essential for robust negation detection.
%


\section{LLM-Based Negation Detection in Political Multimodal Discourse}

To analyze negation in political multimodal content, we employ a prompting-based framework using the multimodal large language model \textbf{Qwen-VL 2.5} \cite{bai:et:al:2025}. 
This model is queried with video-text pairs using a structured prompt designed to assess negation and alignment within political communication.
The framework explicitly targets discourse strategies used to reject, contradict, or oppose claims across modalities.
\paragraph*{Modeling Setup.}
Each input consists of two components: a text excerpt derived from a political video (via transcription or manual captioning) and a structured video representation.
The model is prompted to analyze the modalities independently and jointly, producing structured outputs capturing negation presence, type, confidence, alignment, and political intent.

\paragraph*{Prompt Structure.}
The prompt (see Appendix~\ref{appendix:prompt}) is crafted for political discourse analysis and includes three stages:
\begin{enumerate}
    \item \textbf{Textual Negation Detection:} Identifies explicit and implicit political negation using markers such as “not,” “deny,” or “oppose.”
    It classifies the negation type (e.g., policy rejection, factual contradiction) and assigns a confidence score.
    
    \item \textbf{Visual Negation Detection:} Analyzes the video for multimodal signals of negation such as contradictory imagery, disapproving gestures, or symbolic opposition. 
    Negation is classified into types such as evidentiary contradiction or emotional rejection, with associated confidence and cue annotations.
    
    \item \textbf{Modality Alignment:} Assesses the relationship between textual and visual negation. 
    Four alignment outcomes are possible: \textrm{FULL ALIGNMENT}, \textrm{PARTIAL ALIGNMENT}, \textrm{SINGLE MODALITY}, and \textrm{UNRELATED}. 
    Each output also includes a description and inferred strategic political effect (e.g., reinforcement or audience targeting).
\end{enumerate}

\paragraph*{Output Format.}

The model outputs a structured JSON object encoding all decisions made in the three stages above.
Each output includes both categorical and confidence information, along with supporting cues and key phrases that can be used for further analysis or annotation validation.
This method enables scalable, politically informed annotation of negation phenomena in multimodal discourse, without requiring additional training of the underlying model.

\subsection{LLM-Based Negation Analysis}

\subsubsection{Modality-Specific Expression vs. Cross-Modal Semantics}

Our working hypothesis is that \textit{negation manifests as a modality-specific feature}, observable independently in either text or video. 
However, the \textit{semantic interpretation of negation}, such as whether it constitutes contradiction, rejection, or denial, \textit{emerges primarily through multimodal fusion}, when both streams are aligned and interpreted jointly.

\subsubsection{Negation Frequency in Text and Video}

Negation appears more frequently in text (675 instances) than in video (372 instances), with the majority of data (over 75 percent) containing no detectable negation in either modality.
This supports the idea that \textit{negation is a modality-dependent signal}, more easily detected in linguistic form. 
Its infrequent appearance in video suggests that \textit{visual negation is either semantically subtler or harder to isolate without fusion with textual cues}.
%


\subsubsection{Negation Confidence Score Distribution}

\AMC{Text-based negation confidence clusters tightly around 85, with minimal variance.
%
This uniformity may reflect a calibration bias or fixed threshold, while video confidence shows wider spread, with a 90 median and notable outliers (please refer to \autoref{fig:negation_confidence} for more details).
%
These differences suggest video-based negation is less stable and relies on contextual cues best resolved multimodally.}

\subsubsection{Negation Type Frequencies}

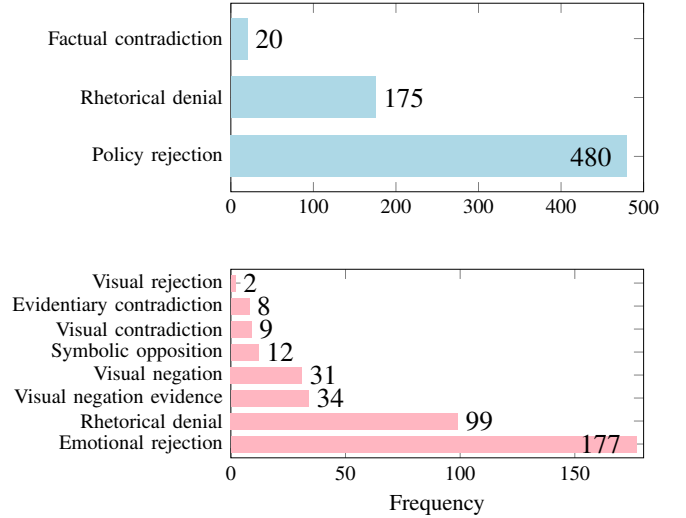
\begin{figure}[t]
\centering
    \resizebox{\linewidth}{!}{\begin{tikzpicture}

\definecolor{darkgrey176}{RGB}{176,176,176}
\definecolor{lightblue}{RGB}{173,216,230}
\definecolor{lightpink}{RGB}{255,182,193}

\begin{groupplot}[
    group style={group size=1 by 2, vertical sep=1.2cm},
    width=0.9\linewidth,
    height=4.5cm
]

\nextgroupplot[
    x grid style={darkgrey176},
    y grid style={darkgrey176},
    ymin=-0.6, ymax=2.6,
    xmin=0, xmax=500,
    ytick={0,1,2},
    yticklabels={
        Policy rejection,
        Rhetorical denial,
        Factual contradiction
    },
    yticklabel style={align=center, font=\small},
    xticklabel style={font=\small},
]

\draw[draw=none, fill=lightblue] (axis cs:0,-0.35) rectangle (axis cs:480,0.35);
\draw[draw=none, fill=lightblue] (axis cs:0,0.65) rectangle (axis cs:175,1.35);
\draw[draw=none, fill=lightblue] (axis cs:0,1.65) rectangle (axis cs:20,2.35);

\node[font=\large, anchor=west, xshift=-1.0cm] at (axis cs:480,0) {480};
\node[font=\large, anchor=west] at (axis cs:175,1) {175};
\node[font=\large, anchor=west] at (axis cs:20,2) {20};

\nextgroupplot[
    x grid style={darkgrey176},
    y grid style={darkgrey176},
    xlabel={Frequency},
    ymin=-0.6, ymax=7.6,
    xmin=0, xmax=180,
    ytick={0,1,2,3,4,5,6,7},
    yticklabels={
        Emotional rejection,
        Rhetorical denial,
        Visual negation evidence,
        Visual negation,
        Symbolic opposition,
        Visual contradiction,
        Evidentiary contradiction,
        Visual rejection
    },
    yticklabel style={align=center, font=\small},
    xticklabel style={font=\small}
]

\draw[draw=none,fill=lightpink] (axis cs:0,-0.35) rectangle (axis cs:177,0.35);
\draw[draw=none,fill=lightpink] (axis cs:0,0.65) rectangle (axis cs:99,1.35);
\draw[draw=none,fill=lightpink] (axis cs:0,1.65) rectangle (axis cs:34,2.35);
\draw[draw=none,fill=lightpink] (axis cs:0,2.65) rectangle (axis cs:31,3.35);
\draw[draw=none,fill=lightpink] (axis cs:0,3.65) rectangle (axis cs:12,4.35);
\draw[draw=none,fill=lightpink] (axis cs:0,4.65) rectangle (axis cs:9,5.35);
\draw[draw=none,fill=lightpink] (axis cs:0,5.65) rectangle (axis cs:8,6.35);
\draw[draw=none,fill=lightpink] (axis cs:0,6.65) rectangle (axis cs:2,7.35);

\node[font=\large, anchor=west, xshift=-1.0cm] at (axis cs:177,0) {177};
\node[font=\large, anchor=west] at (axis cs:99,1) {99};
\node[font=\large, anchor=west] at (axis cs:34,2) {34};
\node[font=\large, anchor=west] at (axis cs:31,3) {31};
\node[font=\large, anchor=west] at (axis cs:12,4) {12};
\node[font=\large, anchor=west] at (axis cs:9,5) {9};
\node[font=\large, anchor=west] at (axis cs:8,6) {8};
\node[font=\large, anchor=west] at (axis cs:2,7) {2};

\end{groupplot}

\end{tikzpicture}}
\caption{Overview of negation types across modalities.}
\label{fig:negation_types_overview}
\end{figure}

The taxonomy of negation types shows that text mainly encodes propositional rejections, like policy or rhetorical denials.
\AMC{In contrast, video conveys broader emotive and symbolic negation, like visual contradiction or emotional rejection.
However, such cases are sparse and hard to interpret alone.}
\AM{This supports the view that negation is modality-specific and gains meaning only through cross-modal fusion.}

\subsubsection{Text-Video Negation Co-occurrence}

\begin{table}[t]
\centering
\renewcommand{\arraystretch}{1.3}
\caption{Co-occurrence between text and video negation}

\begin{tabular}{l|l|c|c|c}
\multicolumn{2}{c}{} & \multicolumn{2}{c}{Video negation} & \\
\cline{3-4}
\multicolumn{2}{c|}{} & Present & Absent & \multicolumn{1}{c}{Total} \\
\cline{2-4}
\multirow{2}{*}{\makecell{Text \\ negation}} 
& Present & 298 & 377 & 675 \\
\cline{2-4}
& Absent & 74 & 2473 & 2547 \\
\cline{2-4}
\multicolumn{1}{c}{} 
& \multicolumn{1}{c}{Total} 
& \multicolumn{1}{c}{372} 
& \multicolumn{1}{c}{2850} 
& \multicolumn{1}{c}{3222} \\
\end{tabular}
\label{tab:co_occurrence}
\end{table}

As shown in \autoref{tab:co_occurrence}, while 298 instances show co-occurrence of negation in both modalities, mismatches are common. There are 377 cases where only text expresses negation and 74 where only video does. 
These results suggest that {the presence of negation is not itself predictive of alignment}, and may reflect {modality-specific encoding of intent}. 
However, when both modalities express negation, the joint presence is more likely to contribute to {semantic-level interpretations} such as contradiction or conflict.
A chi-squared test of independence confirms a strong association between text and video negation ($\chi^2(1) = 888.7$, $p < 0.001$, $\phi = 0.53$)\footnote{(see Appendix~\ref{app:negation_independence} for detailed analysis)}, with the observed co-occurrence (\numprint{298} instances) far exceeding the expected count under independence (77.93).
\subsubsection{Negation vs. Alignment Type}

\begin{table}[h!]
\centering
\begin{tabular}{lcccccc}
\hline
\makecell{Negation \\ Source} & \makecell{ Negation \\ Exists} & FA & PA & SM & U \\
\hline
\multirow{2}{*}{\makecell{Text \\ negation}} 
  & No     & 0  & 74  & 14 & 2165 \\
  & Yes & 47 & 348 & 86 & 2447 \\
\hline
\multirow{2}{*}{\makecell{Video \\ negation}}
    & No & 0 & 23 & 0 & 0 \\
 & Yes & 47 & 348 & 86 & 2447 \\
\hline
\end{tabular}
\caption{Alignment type distribution in negated media. The x-axis shows alignment types (FA: \textsc{FULL-ALIGNMENT}, PA: \textsc{PARTIAL-ALIGNMENT}, SM: \textsc{SINGLE MODALITY}, U: \textsc{UNRELATED}).} 
\end{table}

\textcolor{black}{Text negation is prevalent even in \textsc{UNRELATED} alignments, confirming that {textual negation is often modality-local and independent of cross-modal support}.}
In contrast, video negation occurs exclusively in aligned examples (\textsc{FULL-ALIGNMENT} or \textsc{PARTIAL-ALIGNMENT}), and is completely absent in unrelated or unimodal pairs. 
This provides strong evidence for the hypothesis that {video-based negation is semantically dependent on textual context} and {only becomes interpretable when modalities are fused}.
%

\begin{table}[t]
\centering
\caption{Performance of multihead vs.\ cross-modal attention (30 epochs, 4 heads, LLM-labeled negation)}
\label{tab:attention_results}
\resizebox{\linewidth}{!}{
\begin{tabular}{@{}cccccclc@{}}
\toprule
\multirow{2}{*}{\textbf{\makecell{ Attention \\ Heads }}} & \multirow{2}{*}{\textbf{Type}} & \multicolumn{3}{c}{\textbf{Metrics}} & \multicolumn{2}{c}{\textbf{Input}} & \multirow{2}{*}{\textbf{\makecell{Target \\ LLM-negation}}} \\
\cmidrule(lr){3-5} \cmidrule(lr){6-7}
 & & Acc. & F1 & ROC-AUC & Text & Video & \\ 
\midrule
\multicolumn{8}{c}{\textit{Multihead Attention (Qwen)}} \\
\midrule
8 & SA-Video & 0.8650 & 0.8430 & 0.7337 & & \checkmark & Video-Neg \\
8 & SA-Text  & 0.8650 & 0.8640 & 0.9174 & \checkmark & & Text-Neg \\
\midrule
\multicolumn{8}{c}{\textit{Multihead Attention (Jepa2)}} \\
\midrule
8 & SA-Video & 0.8704 & 0.8684 & 0.7931 & & \checkmark & Video-Neg \\
8 & SA-Text  & 0.8150 & 0.8191 & 0.8829 & \checkmark & & Text-Neg \\ 
\midrule
\multicolumn{8}{c}{\textit{Cross-Modal Attention (Qwen Video + Text)}} \\
\midrule
4 & CM-Video & 0.8525 & 0.8460 & 0.7451 & \checkmark & \checkmark & Video-Neg \\
4 & CM-Text  & 0.8350 & 0.8226 & 0.8545 & \checkmark & \checkmark & Text-Neg \\
\midrule
\multicolumn{8}{c}{\textit{Cross-Modal Attention (Jepa2 Video + Qwen Text)}} \\
\midrule
4 & CM-Video & 0.8862 & 0.8832 & 0.8318 & \checkmark & \checkmark & Video-Neg \\
4 & CM-Text  & 0.8651 & 0.8693 & 0.9320 & \checkmark & \checkmark & Text-Neg \\
\bottomrule
\end{tabular}}
\vspace{0.2cm}
\end{table}

\subsubsection{Human Validation of LLM Annotations}
\label{sec:human_validation}

To assess LLM annotation reliability, we compared Qwen2.5-VL annotations against two generalist human experts (BU: PhD student; MO: postdoc) across three experimental setups: (1) original 3B model with full multimodal input, (2) same 3B model with text only, 
and (3) new 7B model with revised prompt (multimodal).
The two human annotators show near-perfect agreement ($\kappa=1.0$), establishing a reliable reference. 
The original multimodal model yields near-zero agreement with humans ($\kappa=0.07$ with BU; $\kappa=-0.0$ with MO). 
When the same 3B model is given only tweet text, agreement rises dramatically to moderate levels ($\kappa=0.53$ with BU; $\kappa=0.47$ with MO), confirming the LLM's inherent capability for textual negation detection. 
%
%
Scaling to 7B with the same multimodal input improves agreement to $\kappa=0.19$ with BU and $\kappa=0.24$ with MO, but remains well below text-only levels. 
These results demonstrate that visual information systematically degrades LLM negation performance, and simply increasing model size only partially compensates--effective multimodal negation requires specialized fusion architectures like our cross-modal attention framework.

\section{Discussion}
\label{sec:discussion}

This study investigated the detection and representation of negation in multimodal political discourse through a series of progressively complex experiments. 
Our findings reveal a clear trajectory: while negation as a semantic concept is readily identifiable by humans and advanced LLMs, it does not manifest as a strong, separable signal within the latent embeddings of individual modalities when using standard supervised classifiers.

\paragraph{The Embedding-Level Challenge} 
%
Initial experiments using diverse classifiers (linear, tree-based, ensemble) on modality-specific embeddings (video, image, text) yield a key null result.
Despite high nominal accuracy for some models (e.g., $\sim$ \numprint{81}\% for Random Forests), balanced accuracy and ROC AUC scores remain at chance levels ($\sim \numprint{0.5}$). 
This performance was matched by a DummyClassifier, indicating that the negation labels derived from text were {not mapping to linearly or non-linearly separable regions in the video or image embedding spaces.} 
PCA visualizations corroborated this, showing that the primary variance in these embeddings was attributable to modality-specific features rather than to the presence or absence of negation.

\paragraph{The Modality-Alignment Imperative} The RNN-based experiments (Table~\ref{tab:rnn-model-metrics}) provide further subtlety. 
Performance is highest when the modality of the input embedding match the source modality of the label (e.g., text embeddings with text-derived negation labels, photo embeddings with photo-derived labels).
Models trained on cross-modal label pairs (e.g., video embeddings with text-derived labels) perform poorly. 
This suggests that while pretrained \textbf{embeddings are rich representations} of their native modality, they \textbf{do not 
encode a transferable negation signal} that can be recognized using labels from another modality. 
The high performance in matched-modality scenarios may point more towards dataset-specific correlations or labeling artifacts than to a generalizable negation feature.

\paragraph{Cross-Modal Fusion as a Solution} 
The failure of unimodal classification led to the hypothesis that negation is a cross-modal semantic construct. 
The cross-modal attention framework validated this.
By jointly processing text and video embeddings, the model learns intermodal dependencies, allowing it to disambiguate and fuse signals that are weak or confusing in isolation. 
%
The cross-modal attention model outperformed unimodal baselines, with gains up to $+$\numprint{4.53}\% accuracy and $+$\numprint{7.03}\% F1 for text sentiment.
%
\textit{This shows that negation in discourse, such as politics, is often best understood through language-vision interaction: 
\AM{in this context, negation functions as a cross-modal communicative unit}.}

\paragraph{LLM Analysis} 
Prompt-based analysis with Qwen-VL 2.5 provides strong upper-bound validation of our core hypothesis.
%
\AM{The LLM analysis confirms that negation differs by modality: text conveys explicit rejection, while video expresses subtler emotive and symbolic forms.}
Crucially, the alignment analysis revealed a fundamental asymmetry: textual negation often stands alone, whereas visual negation was only identified in the context of aligned text. This provides strong evidence that visual negation is semantically dependent on linguistic context for its interpretation. 
%
The success of both self-attention and cross-modal classifiers (Table~\ref{tab:attention_results}) shows that when negation is represented via high-level semantic concepts (as in LLM-generated labels), it becomes a learnable target, \textit{especially for models with multimodal fusion}.
%
\paragraph{Classifier Capacity vs. Signal Absence.}
A potential concern is whether our negative findings (classifiers performing at chance) stem from weak classifiers rather than the absence of negation signals in embeddings. Several observations argue against this interpretation. 
First, our classifier suite \autoref{tab:negation_perf} included 26 diverse models spanning multiple paradigms (linear, tree-based, ensemble, nearest neighbors), all yielding chance-level balanced accuracy. 
Second, sophisticated models (XGBClassifier, Random Forest) performed nearly identically to a stratified DummyClassifier, indicating the issue is signal absence, not classifier capacity. 
Third, our embeddings are static, pooled representations (2048-dim); Transformer-based classifiers would operate on the same feature space without access to sequential information that could theoretically aid separation. 
While sequence-aware models might capture different patterns, they cannot create separable negation signals where none exist in the pooled representations. 
This supports our conclusion that negation is not linearly separable in standard VLM embeddings.
\section{Conclusion}
\label{sec:conclusion}

We investigated negation detection in multimodal political video and draw three main conclusions:
\textit{Negation is not an embedding-level signal:} 
\AMC{Supervised classifiers fail to detect negation across modalities because current pretrained embeddings encode weak or entangled modality-specific signals.}
%
%
Negation is a cross-modal phenomenon whose expression and interpretation often rely on the alignment of textual and visual cues.
To model this effectively and surpass unimodal baselines, architectures that support multimodal fusion such as cross-modal attention are essential.
%
\textit{LLMs provide a viable framework for high-level negation annotation.} 
\AMC{Thus, by leveraging \AM{VLM} reasoning, we generate structured negation and alignment labels that capture pragmatic meaning and supervise models bridging low-level embeddings and high-level rhetoric.}
%
\AMC{Future work focuses on end-to-end architectures to learn cross-modal negation semantics from raw data, potentially using LLM-derived annotations as weak supervision. 
Exploring a finer-grained taxonomy of negation types and their cross-modal interaction patterns could offer insights into multimodal discourse.}

\section{Limitations \& Ethical Considerations}
\paragraph{Limitations}Our study has several limitations that warrant consideration. 
First, the LLM-based annotations, while scalable, may inherit biases from Qwen-VL 2.5's training data and could miss nuanced negation forms.
Second, we primarily evaluated Qwen2.5-VL and JEPA2 embeddings; other architectures may encode negation differently. 
Third, the LLM's confidence scores showed poor calibration-text negation confidence was uniformly \numprint{85}, limiting their reliability for uncertainty-aware applications.
Fourth, our focus on political discourse may limit generalizability to other domains where negation functions differently. 
Finally, the computational demands of cross-modal attention may hinder accessibility.
Despite these constraints, our work establishes foundational insights into multimodal negation and points toward future research in calibrated multimodal LLMs, broader model evaluations, and ethical applications.
\paragraph{Ethical Considerations} Our study analyzes German political discourse using automated annotation via Qwen2.5-VL, which may not be fully optimized for German language and cultural nuances, potentially introducing systematic biases.
The use of negation detection in political content could be misapplied for surveillance or to discredit opposing views. 
%
We recommend human validation of automated labels and transparency in deployment to mitigate these risks.


\bibliographystyle{IEEEtran}
\bibliography{bibtex}

\appendix


\section{LLM-Based Negation analysis}
\label{appdenix:llm_negation_analysis}

This appendix presents methodological and analytical details supporting the LLM based negation detection framework.

It includes the structured prompt used to obtain negation and alignment annotations from Qwen2.5VL, along with an analysis of confidence scores across modalities.

\subsection{Prompt Used for Political Negation Detection}
\label{appendix:prompt}
The following system prompt was used to elicit structured negation and alignment annotations from the Qwen-VL 2.5 model. It is tailored for political discourse across text and video.
\begin{lstlisting}[language=JSON, caption={Prompt for political negation detection}, label={lst:negation-prompt}]
You are an expert in political discourse analysis specializing in negation detection across media formats.  Analyze the text and video to determine: 
1. Whether negation exists in EACH modality separately 
2. If negations align between text and video 
Inputs Provided: 
- Political text excerpt: [TEXT ADDED HERE]
- Video : Attached 
Analysis Framework 
A. TEXT NEGATION VERIFICATION: 
   [ ] Check for explicit negation markers: 
       * Direct negatives (not, never, no) 
       * Implicit denial terms (deny, reject, oppose) 
   [ ] Identify political negation type: 
       * Policy rejection 
       * Factual contradiction 
       * Rhetorical denial 
       * Opposition framing 
   [ ] Score confidence: 0-100 
B. VIDEO NEGATION VERIFICATION: 
   [ ] Check for visual negation evidence: 
       * Contradictory imagery (e.g., prosperity visuals during austerity claims) 
       * Affective cues (disapproval gestures, head-shaking) 
       * Contextual dissonance (incongruous setting/symbols) 
       * Absence portrayal (missing expected elements) 
   [ ] Classify visual negation type: 
       * Evidentiary contradiction 
       * Emotional rejection 
       * Symbolic opposition 
   [ ] Score confidence: 0-100 
C. MODALITY ALIGNMENT ASSESSMENT: 
   [ ] Determine negation correspondence: 
       * FULL ALIGNMENT: Both modalities contain matching negation 
       * PARTIAL ALIGNMENT: One modality negates what the other affirms 
       * UNRELATED: No meaningful connection 
       * SINGLE-MODALITY: Negation only present in one channel 
   [ ] Identify political intent: 
       * Reinforcement strategy 
       * Doubt creation 
       * Audience targeting 
Output Requirements, Strict JSON format: 
{"text_negation": { 
    "exists": boolean, "type": "string",  "confidence": integer, "key_phrases": ["string"] }, 
  "video_negation": { 
    "exists": boolean, "type": "string", 
    "confidence": integer, "visual_cues": ["string"] }, 
  "alignment": { 
    "type": "FULL-ALIGNMENT / PARTIAL-ALIGNMENT / UNRELATED / SINGLE-MODALITY", 
    "description": "string", "strategic_effect": "string" } } 
Political Analysis Focus 
- Prioritize policy/claim negation over general negativity 
- Note power dynamics in denial (e.g., institution vs individual) 
- Flag strategic ambiguity techniques 
- Identify visual framing of opposition 
Example: 
Text: "The administration claims economic growth" 
Video: Graphs showing declining GDP with expert head-shaking 
Output:  {"text_negation": {"exists": false, ...}, "video_negation": {"exists": true, "type": "Evidentiary contradiction", ...},  "alignment": {"type": "SINGLE-MODALITY", ...} }
\end{lstlisting}

\subsection{Negation Confidence Score}

We present the distribution of confidence scores assigned by the LLM for its negation detections across text and video modalities. 

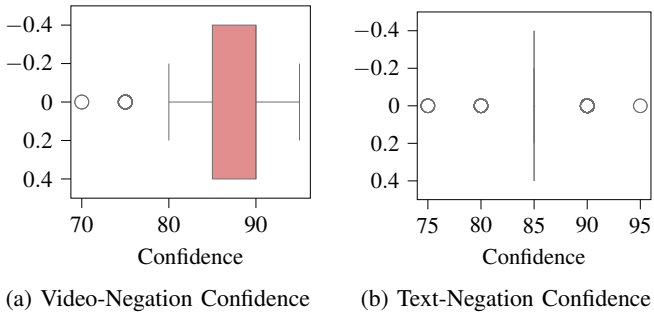
\begin{figure}[ht]
\centering
\begin{subfigure}[b]{0.48\columnwidth}
    \resizebox{\linewidth}{!}{
\begin{tikzpicture}

\definecolor{darkgrey176}{RGB}{176,176,176}
\definecolor{darksalmon226142142}{RGB}{226,142,142}
\definecolor{dimgrey110}{RGB}{110,110,110}

\begin{axis}[
tick align=outside,
tick pos=left,
x grid style={darkgrey176},
xlabel={Confidence},
xmin=68.75, xmax=96.25,
xtick style={color=black},
y dir=reverse,
y grid style={darkgrey176},
ymin=-0.5, ymax=0.5,
ytick style={color=black},
height=4.5cm
]
\path [draw=dimgrey110, fill=darksalmon226142142]
(axis cs:85,-0.4)
--(axis cs:85,0.4)
--(axis cs:90,0.4)
--(axis cs:90,-0.4)
--(axis cs:85,-0.4)
--cycle;
\addplot [dimgrey110]
table {%
85 0
80 0
};
\addplot [dimgrey110]
table {%
90 0
95 0
};
\addplot [dimgrey110]
table {%
80 -0.2
80 0.2
};
\addplot [dimgrey110]
table {%
95 -0.2
95 0.2
};
\addplot [black, mark=o, mark size=3, mark options={solid,fill opacity=0,draw=dimgrey110}, only marks]
table {%
75 0
75 0
75 0
75 0
75 0
75 0
75 0
75 0
75 0
75 0
75 0
75 0
75 0
75 0
75 0
70 0
75 0
75 0
75 0
75 0
75 0
75 0
75 0
75 0
70 0
75 0
};
\addplot [dimgrey110]
table {%
90 -0.4
90 0.4
};
\end{axis}

\end{tikzpicture}}
    \caption{Video-Negation Confidence}
    \label{fig:video_negation_confidence}
\end{subfigure}
\hfill
\begin{subfigure}[b]{0.48\columnwidth}
    \resizebox{\linewidth}{!}{
\begin{tikzpicture}

\definecolor{darkgrey176}{RGB}{176,176,176}
\definecolor{dimgrey114}{RGB}{114,114,114}
\definecolor{lightgreen155226155}{RGB}{155,226,155}

\begin{axis}[
tick align=outside,
tick pos=left,
x grid style={darkgrey176},
xlabel={Confidence},
xmin=74, xmax=96,
xtick style={color=black},
y dir=reverse,
y grid style={darkgrey176},
ymin=-0.5, ymax=0.5,
ytick style={color=black},
height=4.5cm
]
\path [draw=dimgrey114, fill=lightgreen155226155]
(axis cs:85,-0.4)
--(axis cs:85,0.4)
--(axis cs:85,0.4)
--(axis cs:85,-0.4)
--(axis cs:85,-0.4)
--cycle;
\addplot [dimgrey114]
table {%
85 0
85 0
};
\addplot [dimgrey114]
table {%
85 0
85 0
};
\addplot [dimgrey114]
table {%
85 -0.2
85 0.2
};
\addplot [dimgrey114]
table {%
85 -0.2
85 0.2
};
\addplot [black, mark=o, mark size=3, mark options={solid,fill opacity=0,draw=dimgrey114}, only marks]
table {%
80 0
80 0
80 0
80 0
80 0
80 0
80 0
80 0
75 0
80 0
80 0
80 0
75 0
75 0
80 0
80 0
75 0
75 0
80 0
75 0
75 0
80 0
80 0
80 0
75 0
80 0
80 0
80 0
80 0
80 0
75 0
80 0
90 0
90 0
90 0
90 0
90 0
90 0
90 0
90 0
90 0
90 0
90 0
90 0
90 0
90 0
90 0
90 0
90 0
90 0
90 0
90 0
90 0
90 0
90 0
90 0
90 0
90 0
90 0
90 0
90 0
90 0
90 0
90 0
90 0
90 0
90 0
90 0
90 0
90 0
90 0
90 0
90 0
90 0
90 0
90 0
90 0
90 0
90 0
90 0
90 0
90 0
90 0
90 0
90 0
90 0
90 0
90 0
90 0
90 0
90 0
90 0
90 0
90 0
90 0
90 0
90 0
90 0
90 0
90 0
90 0
90 0
90 0
90 0
90 0
90 0
90 0
90 0
90 0
90 0
90 0
90 0
90 0
90 0
90 0
90 0
90 0
90 0
90 0
90 0
90 0
95 0
90 0
95 0
90 0
90 0
90 0
90 0
90 0
90 0
90 0
90 0
90 0
90 0
90 0
90 0
90 0
90 0
90 0
90 0
90 0
90 0
90 0
90 0
90 0
90 0
90 0
90 0
90 0
90 0
90 0
90 0
90 0
90 0
90 0
90 0
90 0
90 0
90 0
90 0
90 0
90 0
90 0
90 0
90 0
90 0
90 0
90 0
90 0
90 0
90 0
90 0
90 0
90 0
90 0
90 0
90 0
90 0
90 0
90 0
90 0
90 0
90 0
90 0
90 0
90 0
90 0
90 0
90 0
90 0
90 0
90 0
90 0
90 0
90 0
90 0
90 0
90 0
};
\addplot [dimgrey114]
table {%
85 -0.4
85 0.4
};
\end{axis}

\end{tikzpicture}}
    \caption{Text-Negation Confidence}
    \label{fig:text_negation_confidence}
\end{subfigure}
\caption{Overview of negation confidence across modalities.}
\label{fig:negation_confidence}
\end{figure}

\subsection{Statistical Analysis of Negation Co-occurrence}
\label{app:negation_independence}
To statistically validate the relationship between text and video negation
observed in Table~\ref{tab:co_occurrence}, we conducted a chi squared test of
independence. The expected frequencies under the null hypothesis are
\begin{equation}
\begin{aligned}
E_{11} &= \frac{675 \times 372}{3222} = 77.93, \hspace{0.2cm}
E_{12} &= \frac{675 \times 2850}{3222} = 597.07, \\
E_{21} &= \frac{2547 \times 372}{3222} = 294.07, \hspace{0.2cm}
E_{22} &= \frac{2547 \times 2850}{3222} = 2276.82.
\end{aligned}
\label{eq:expected}
\end{equation}

The chi squared statistic is computed as
\begin{equation}
\chi^2 = \sum_{i,j} \frac{(O_{ij} - E_{ij})^2}{E_{ij}},
\label{eq:chi_def}
\end{equation}
which yields
\begin{equation}
\chi^2 = 621.465 + 81.144 + 164.691 + 21.499 \approx 888.7.
\label{eq:chi_value}
\end{equation}

The corresponding phi coefficient is given by
\begin{equation}
\phi = \sqrt{\frac{\chi^2}{N}} = \sqrt{\frac{888.7}{3222}} \approx 0.53.
\label{eq:phi}
\end{equation}

With one degree of freedom, this result yields $p < 0.001$, indicating a
statistically significant association between text and video negation.
The phi coefficient confirms a moderate to strong positive correlation.

\section{Algorithm Paradigms for classifications}

This appendix show the algorithms used for our baseline, the table \ref{tab:negation_perf} represents the classifier taxonomy, and the  results for the baseline.


\begin{table}[t]
\centering
\caption{Classification performance on media embeddings for negation detection.}
\resizebox{\columnwidth}{!}{%
\begin{tabular}{lllllll}
\toprule
Taxonomy Type &Model & Accuracy & \makecell{Balanced \\ Accuracy} & ROC AUC & F1 Score & \makecell{Time \\ Taken} \\
\midrule
Naive Bayes & GaussianNB & 0.577 & 0.542 & 0.542 & 0.621 & 0.610 \\
Naive Bayes & BernoulliNB & 0.570 & 0.537 & 0.537 & 0.616 & 0.710 \\
Tree-Based & DecisionTreeClassifier & 0.704 & 0.538 & 0.538 & 0.709 & 48.307 \\
Tree-Based & ExtraTreesClassifier & 0.816 & 0.523 & 0.523 & 0.745 & 10.851 \\
Tree-Based &ExtraTreeClassifier & 0.698 & 0.513 & 0.513 & 0.699 & 0.570 \\
Ensemble & XGBClassifier & 0.812 & 0.530 & 0.530 & 0.749 & 6.931 \\
Ensemble &BaggingClassifier & 0.801 & 0.529 & 0.529 & 0.746 & 402.606 \\
Ensemble &RandomForestClassifier & 0.815 & 0.525 & 0.525 & 0.746 & 56.116 \\
Ensemble &LGBMClassifier & 0.814 & 0.520 & 0.520 & 0.742 & 2.040 \\
Ensemble &CalibratedClassifierCV & 0.810 & 0.500 & 0.500 & 0.724 & 131.109 \\
Ensemble &AdaBoostClassifier & 0.810 & 0.500 & 0.500 & 0.724 & 67.679 \\
Nearest Neighbors & NearestCentroid & 0.581 & 0.537 & 0.537 & 0.624 & 0.564 \\
Nearest Neighbors & KNeighborsClassifier & 0.786 & 0.523 & 0.523 & 0.739 & 0.598 \\
Linear Models & Perceptron & 0.695 & 0.512 & 0.512 & 0.697 & 0.767 \\
Linear Models & LogisticRegression & 0.745 & 0.512 & 0.512 & 0.719 & 0.687 \\
Linear Models &LinearSVC & 0.743 & 0.507 & 0.507 & 0.717 & 31.834 \\
Linear Models &PassiveAggressiveClassifier & 0.703 & 0.507 & 0.507 & 0.699 & 1.045 \\
Linear Models & SGDClassifier & 0.727 & 0.504 & 0.504 & 0.709 & 3.869 \\
Linear Models &RidgeClassifier & 0.773 & 0.503 & 0.503 & 0.723 & 0.699 \\
Linear Models & RidgeClassifierCV & 0.774 & 0.502 & 0.502 & 0.723 & 3.892 \\
Linear Models & SVC & 0.809 & 0.502 & 0.502 & 0.726 & 147.002 \\

Discriminant Analysis& QuadraticDiscriminantAnalysis & 0.811 & 0.528 & 0.528 & 0.748 & 5.091 \\
Discriminant Analysis& LinearDiscriminantAnalysis & 0.741 & 0.509 & 0.509 & 0.717 & 4.590 \\

Semi-Supervised& LabelSpreading & 0.816 & 0.523 & 0.523 & 0.745 & 2.355 \\
Semi-Supervised& LabelPropagation & 0.816 & 0.523 & 0.523 & 0.745 & 2.000 \\

Baseline & DummyClassifier & 0.810 & 0.500 & 0.500 & 0.724 & 0.452 \\
\bottomrule
\end{tabular}%
}
\label{tab:negation_perf}
\end{table}

\section{Principal Component Analysis (PCA) comparison between media embeddings and text-embeddings}

\begin{figure}[ht]
    \centering
    \includegraphics[width=1\linewidth, height=0.5\linewidth]{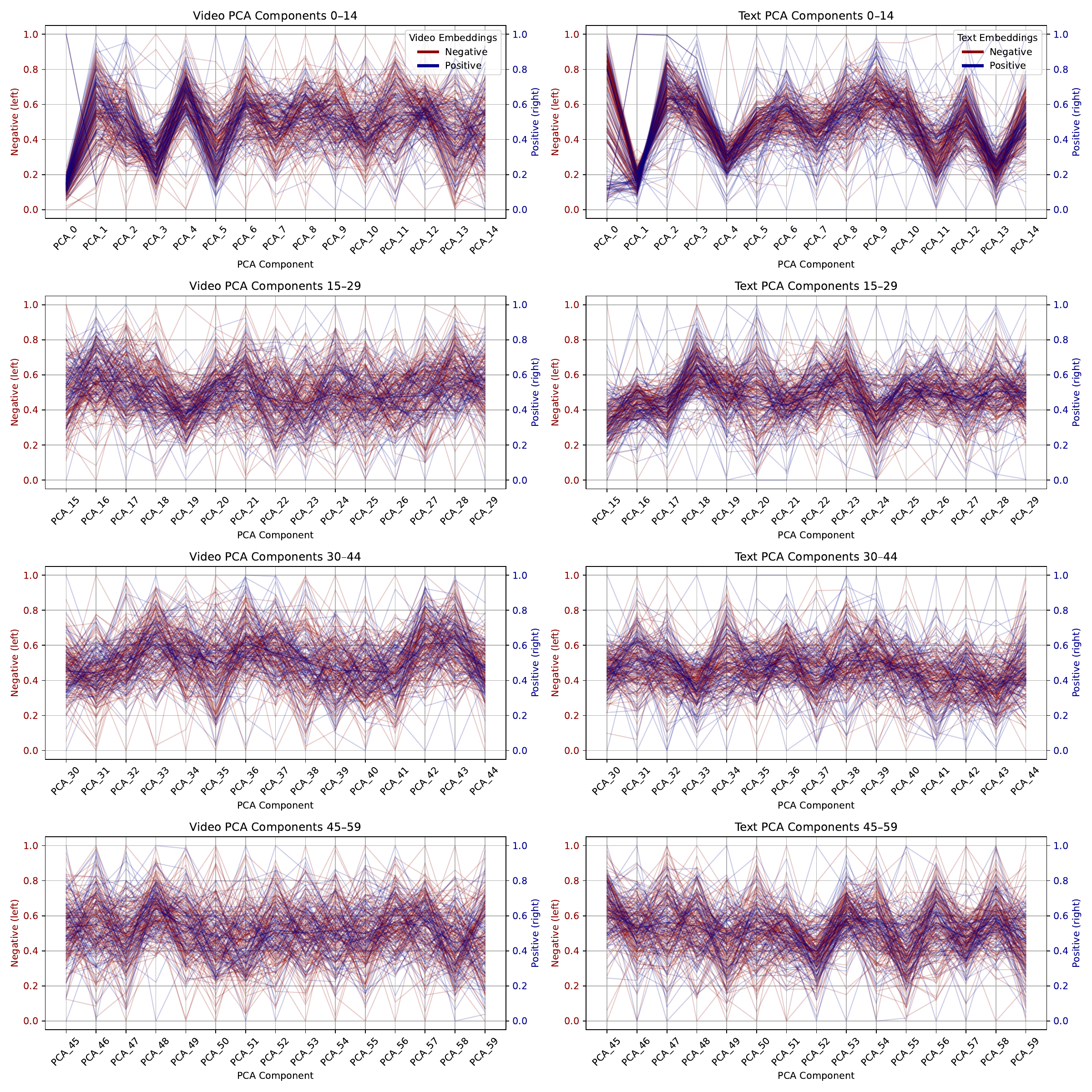}
    \caption{Parallel Coordinates }
    \label{fig:parallel_coordinates}
\end{figure}

This appendix presents the complete PCA visualizations referenced in \autoref{subsection:embeddings_text_label} of the main text. 
The parallel coordinate (\autoref{fig:parallel_coordinates}) plots below illustrate the first 50 principal components of video and text embeddings, colored by their negation labels. 
These visualizations reveal that the primary variance in embeddings is attributable to modality-specific features rather than negation signals, providing further insight into the classification challenges discussed in the main paper.

\end{document}